\documentclass{article} % For LaTeX2e
\usepackage{iclr2027_conference,times}

\usepackage{hyperref}
\usepackage{url}

\usepackage{graphicx}
\usepackage{dblfloatfix}
\usepackage{algorithmic}
\usepackage{booktabs}
\usepackage{multirow}
\usepackage{makecell}
\usepackage{colortbl}
\usepackage[dvipsnames,svgnames,x11names]{xcolor}
\usepackage{amsmath}
\usepackage{amssymb}
\usepackage{algorithm}
\usepackage{algorithmic}
\usepackage{xcolor}
\usepackage{enumitem}
\definecolor{rosebg}{RGB}{248,220,225}
\definecolor{rosefg}{RGB}{180,55,75}
\definecolor{bluegreybg}{RGB}{225,232,240}
\definecolor{bluegreyfg}{RGB}{80,95,115}
\usepackage{wrapfig}
\usepackage{capt-of}
\usepackage{subcaption}
\usepackage{multicol}
\usepackage[most]{tcolorbox}
\definecolor{PPOFamily}{HTML}{EBF5FF}
\definecolor{GDPOFamily}{HTML}{DCECF7}
\definecolor{WebShopBlue}{HTML}{EAF4FB}
\definecolor{WebShopGray}{RGB}{110,110,110}

\iclrfinalcopy

\title{C\ensuremath{^3}: Rethinking Process Supervision via\\ Confidence-Guided Credit Correction in\\ Multi-Turn Reinforcement Learning}

\author{
\textbf{
Ai Jian$^{1,*}$ \quad
Haoran Pan$^{2,*}$ \quad
Eryu Guo$^{1}$ \quad
Haojin Yang$^{2}$ \quad
Jinghan He$^{3}$ 
}
\\
\textbf{
Xiaoyun Zhang$^{4}$ \quad
Jiangbo Pei$^{1}$ \quad
Peiyu Liu$^{5,\dagger}$
}
\\
$^{1}$ Beijing University of Posts and Telecommunications \\
$^{2}$ Peking University \\
$^{3}$ Institute of Automation, Chinese Academy of Sciences \\
$^{4}$ University of Chinese Academy of Sciences \\
$^{5}$ University of International Business and Economics \\
$^{*}$ Equal contribution \qquad
$^{\dagger}$ Corresponding author \\
\texttt{jianai@bupt.edu.cn} \qquad
\texttt{liupeyustu@163.com}
}

\newcommand{\method}{C\ensuremath{^3}}

\newcommand{\meanstd}[2]{#1{\scriptsize$\mathbin{\pm}$#2}}

\begin{document}
\maketitle

\begin{abstract}
Large language models are increasingly trained as multi-turn agents that interact with external tools. However, outcome-only reinforcement learning can produce a form of \emph{misguided reinforcement}: even on successful trajectories, undesirable local behaviors may receive positive token-level advantages and be favored by the policy update. 
To address this issue, we propose \emph{Confidence-Guided Credit Correction (C$^3$)}, a plug-in method for local credit correction. C$^3$ identifies spans corresponding to undesirable behaviors and selectively corrects positive credit after advantage estimation, with the correction strength adaptively controlled by the behavior policy's confidence in the corresponding behavior, while leaving the task reward and original credit estimation unchanged. 
Across multiple multi-turn benchmarks, C$^3$ improves performance under both {critic-based and critic-free reinforcement learning methods}. Injecting the same behavioral signals as turn-level or outcome rewards does not consistently improve performance. Our results show that local credit correction provides an effective form of process supervision for outcome-driven RL.
\end{abstract}
\vspace{-10pt}

\section{Introduction}

Multi-turn agentic reinforcement learning has emerged as an effective paradigm for training language agents on complex tasks that require long-horizon reasoning and iterative tool interaction~\citep{deepresearcher,sweet_rl,trustsql}.
A common approach optimizes agents with outcome-level rewards that evaluate the success of an entire interaction trajectory~\citep{search_r1}.
However, outcome-level supervision provides only sparse and coarse-grained feedback for intermediate behaviors, making it difficult to assign appropriate credit across long-horizon interactions~\citep{stepsearch,igpo,hisr}.
To provide denser supervision for intermediate behaviors, recent work introduces process-level rewards and other process-level signals~\citep{steppo,lotapo,factor}.

\begin{figure*}[t]
	\centering
	\begin{subfigure}[t]{0.412\textwidth}
		\centering
		\includegraphics[width=\linewidth]{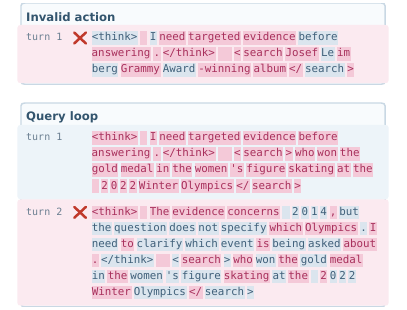}
		\caption{Mixed credit in undesirable behaviors.}
		\label{fig:case}
	\end{subfigure}
	\hfill
	\begin{subfigure}[t]{0.558\textwidth}
		\centering
		\includegraphics[width=\linewidth]{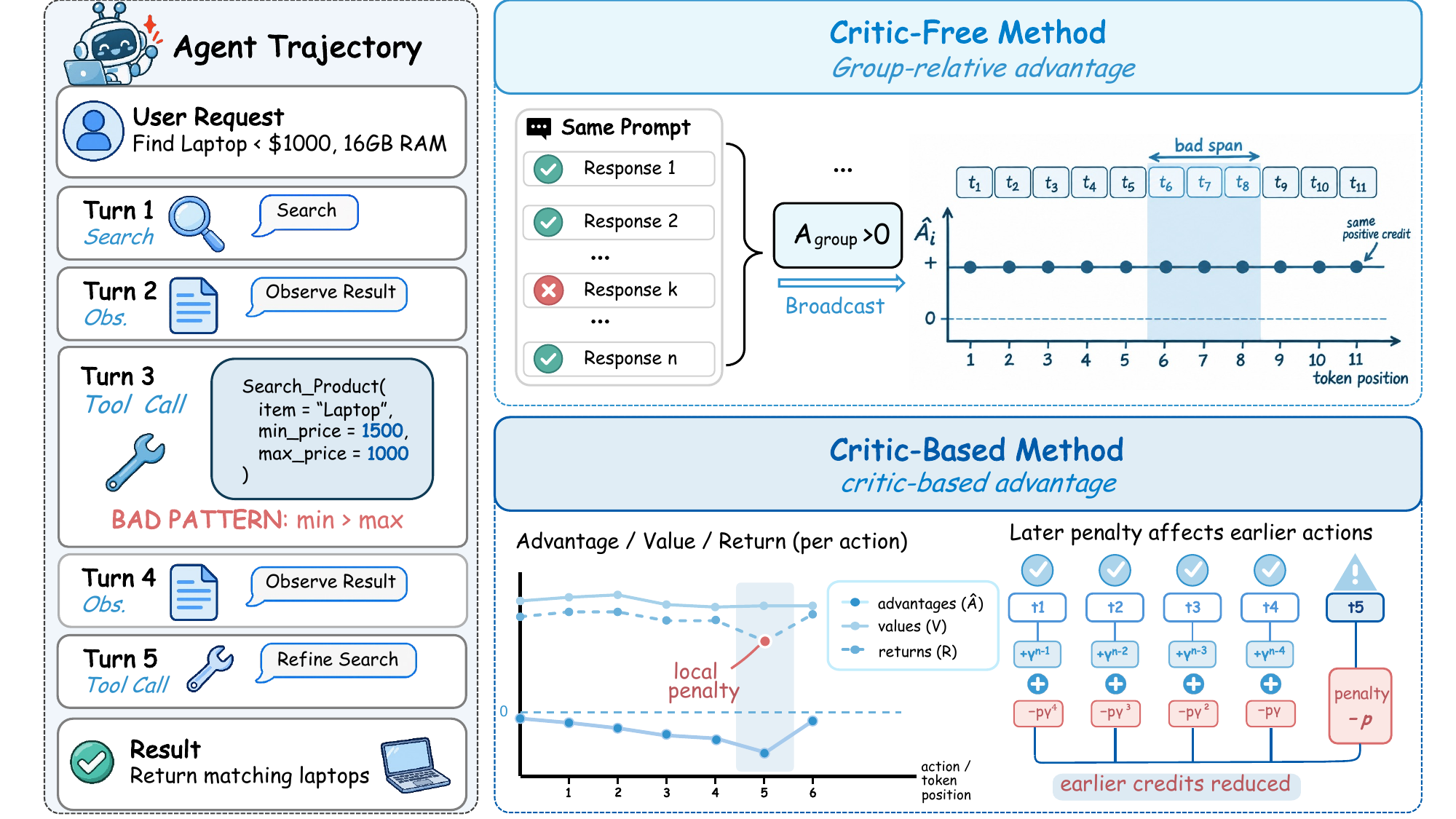}
		\caption{Credit construction across optimization paradigms.}
		\label{fig:intro-concept}
	\end{subfigure}
	\caption{{Misguided reinforcement under outcome-only training.}
		(a)~Incorrect tool use and redundant search receive mixed token-level
		credit, with positive and negative advantages shown in \colorbox{rosebg}{\textcolor{rosefg}{rose}} and \colorbox{bluegreybg}{\textcolor{bluegreyfg}{blue-grey}}.
		(b)~Token-level credit construction in critic-based and critic-free
		optimization.}
	\label{fig:intro}
\end{figure*}
\textbf{Our Key Observation.}
We inspect representative training trajectories under outcome-only training and find that undesirable local behaviors can still receive positive token-level advantages. We identify this phenomenon as a form of \textbf{{misguided reinforcement}}. More precisely, it is a form of local credit misalignment, where the token-level advantage assigned to an undesirable behavior is inconsistent with its process-level quality.
This creates a fundamental mismatch between {\textit{what behavior should be discouraged}} and {\textit{what the policy update favors}}.
Figure~\ref{fig:intro}(a) shows two representative undesirable behaviors observed. In the invalid tool call, a missing closing tag for \texttt{<search>} causes the call to be malformed, yet the preceding unclosed tokens receive positive advantages while the subsequently closed tokens receive negative advantages. In the redundant search, the same query is issued in consecutive turns, but its tokens also receive mixed advantages.
{{Such misalignment reflects the challenge of assigning localized credit from trajectory-level outcomes.}}

\textbf{Why Existing Credit Construction Falls Short.}
Figure~\ref{fig:intro}(b) illustrates how token-level credit is constructed under two common optimization paradigms. In {{critic-based}} reinforcement learning, token-level advantages are estimated through returns and value functions, so a process-level penalty can propagate to tokens beyond the behavior that triggered it. In {{critic-free}} methods, relative advantages are derived directly from trajectory-level outcomes, causing local behaviors to inherit credit from the overall trajectory. 
%Thus, process-level supervision does not inherently provide localized credit correction, \textbf{\textit{motivating the need to correct the credit of undesirable behaviors while preserving the credit of unrelated behaviors.}}
Thus, the two paradigms differ in how credit is constructed, but share a more fundamental
challenge:

{\emph{How can process-level supervision correct a localized undesirable behavior without altering the credit assigned to unrelated behaviors, thereby avoiding credit misalignment?}}

To address this challenge, we propose \textbf{C$^3$ (Confidence-Guided Credit Correction)}, a lightweight process-level credit correction module that directly operates on token-level advantages. 
Given undesirable behaviors localized by external process-level supervision, C$^3$ selectively corrects positive advantages within the corresponding spans, with the correction magnitude adapting to the model's confidence in the corresponding actions. 
C$^3$ introduces no additional process rewards and does not modify the underlying policy optimization objective, making it a {{plug-in module}} applicable to both {{critic-based and critic-free reinforcement learning paradigms}}.

Our contributions are summarized as follows:
\begin{itemize}
	
	\item We identify and characterize misguided reinforcement, where undesirable local behaviors can receive positive policy advantages under outcome-level training, creating a mismatch between behavioral quality and the direction of policy updates.
	
	\item We propose C$^3$, a plug-and-play confidence-guided credit correction framework that rectifies misaligned positive credit within detector-localized undesirable behavior spans without altering task rewards or the underlying RL objective.
	
	\item We evaluate C$^3$ across different policy optimization paradigms and multi-turn agent benchmarks. Results show that C$^3$ {{reduces positive advantages on detected undesirable-behavior spans, decreases their occurrence, and improves overall agent performance}}.
	
\end{itemize}

\section{Related Work}

\textbf{Reinforcement Learning for LLM Agents.}
Language agents are commonly optimized with critic-based methods such as
Proximal Policy Optimization (PPO)~\citep{ppo,ouyang2022training} or
group-relative objectives such as Group Relative Policy Optimization (GRPO)~\citep{grpo}.
They support retrieval~\citep{search_r1} and long-horizon research or collaboration
~\citep{deepresearcher,sweet_rl}, but typically learn from trajectory-level outcomes.
Consequently, an otherwise successful trajectory may still reinforce undesirable intermediate behaviors.
Unlike approaches that modify the overall policy optimization process, \method preserves the underlying objective and instead performs localized credit correction on undesirable behavior spans.

\textbf{Credit Assignment for Multi-Turn Agents.}
Credit assignment asks how trajectory-level outcome credit should be attributed to intermediate actions.
Hierarchical methods allocate credit across turns~\citep{archer,hiper}, while turn- or step-level estimators align advantages with action boundaries
~\citep{turn_ppo,steppo}.
Action-to-token allocation further refines the mapping from outcomes to local advantages~\citep{factor}, and retrospective methods estimate each turn's contribution~\citep{murphy,siop,gigpo,t2po}.
\method{} leaves this outcome-to-advantage mapping intact and operates on the resulting token-level advantages.
While these approaches refine how outcome credit is distributed,
\method{} instead focuses on whether the resulting local credit is
{{consistent with the underlying behavior}}.

\textbf{Process Supervision for LLM Agents.}
Process supervision instead asks whether an intermediate behavior is desirable.
Process reward models score individual steps~\citep{lightman2023verify}; search
agents use step-wise rewards~\citep{stepsearch}, information gain~\citep{igpo}, or
hindsight signals~\citep{hisr}.
Judges and self-distillation provide local assessments~\citep{cwgrpo,gear}, while
counterfactual attribution derives feedback from alternative rollouts
~\citep{lotapo}.
These signals are typically introduced as rewards or advantage shaping.
Rather than introducing additional process rewards, \method{} uses localized
process feedback to {{correct positive advantages assigned to undesirable-behavior spans within the existing token-level advantage}}.

\section{Preliminary}
\label{sec:preliminaries}

We consider a multi-turn reinforcement learning setting where an LLM agent
interacts with an environment to solve a task. Given an input $x$, the agent
follows a policy $\pi_\theta$ to generate an action $a_t$ at interaction turn
$t$, receives an environment observation $o_t$, and eventually produces a
final answer $y$. Over $T$ interaction turns, a complete interaction is
represented as
\begin{equation}
	\tau=(x,a_1,o_1,\ldots,a_T,o_T,y).
	\label{eq:trajectory}
\end{equation}

Each action at turn $t$ is represented by an ordered token sequence:
\begin{equation}
	a_t=(z_i)_{i\in I_t}.
	\label{eq:tokenization}
\end{equation}
Here, $I_t$ denotes the contiguous token span of action $a_t$. We index all
generated tokens globally by $i$ and denote the decoding context of token
$z_i$ by $s_i$.

% During optimization, trajectories are collected under the behavior policy
% $\pi_{\theta_k}$ at update $k$, while $\pi_\theta$ denotes the current policy
% being optimized. A base reinforcement learning algorithm assigns each
% generated token a token-level advantage $\hat A_i$, obtained from either
% critic-based or group-relative credit construction.

% \textit{\textbf{The central challenge is that policy optimization operates on token-level advantages, whereas supervision is often available only at the trajectory level}}. 
% C$^3$ operates directly on the resulting token-level advantages while
% leaving the underlying reward function and credit estimator unchanged.
{During optimization, trajectories are collected under the behavior policy
$\pi_{\theta_k}$ at update $k$, while $\pi_\theta$ denotes the current policy
being optimized. A base reinforcement learning algorithm assigns each
generated token a token-level advantage $\hat A_i$, obtained from either
critic-based or group-relative credit estimation. The policy update is
therefore driven by
\begin{equation}
    \nabla_\theta J(\theta)
    \propto
    \mathbb{E}_{\tau\sim\pi_{\theta_k}}
    \left[
        \sum_i
        \hat A_i
        \nabla_\theta \log \pi_\theta(z_i\mid s_i)
    \right].
    \label{eq:policy_gradient}
\end{equation}
Accordingly, tokens with $\hat A_i>0$ are reinforced by the policy update,
whereas those with $\hat A_i<0$ are suppressed.}

\paragraph{Our goal.}
{Generally, most positive-advantage tokens correspond to behaviors that are consistent with successful task completion. However, a small subset of tokens may exhibit undesirable behaviors yet receive positive advantages, causing them to be reinforced by the base RL objective. C$^3$ targets these sparse misassignments by selectively calibrating their advantages, while leaving the remaining tokens, the reward function, and the base advantage estimator unchanged.}

\section{Methodology}
\label{sec:methodology}

\begin{figure*}[!h]
	\centering
	\includegraphics[width=\textwidth]{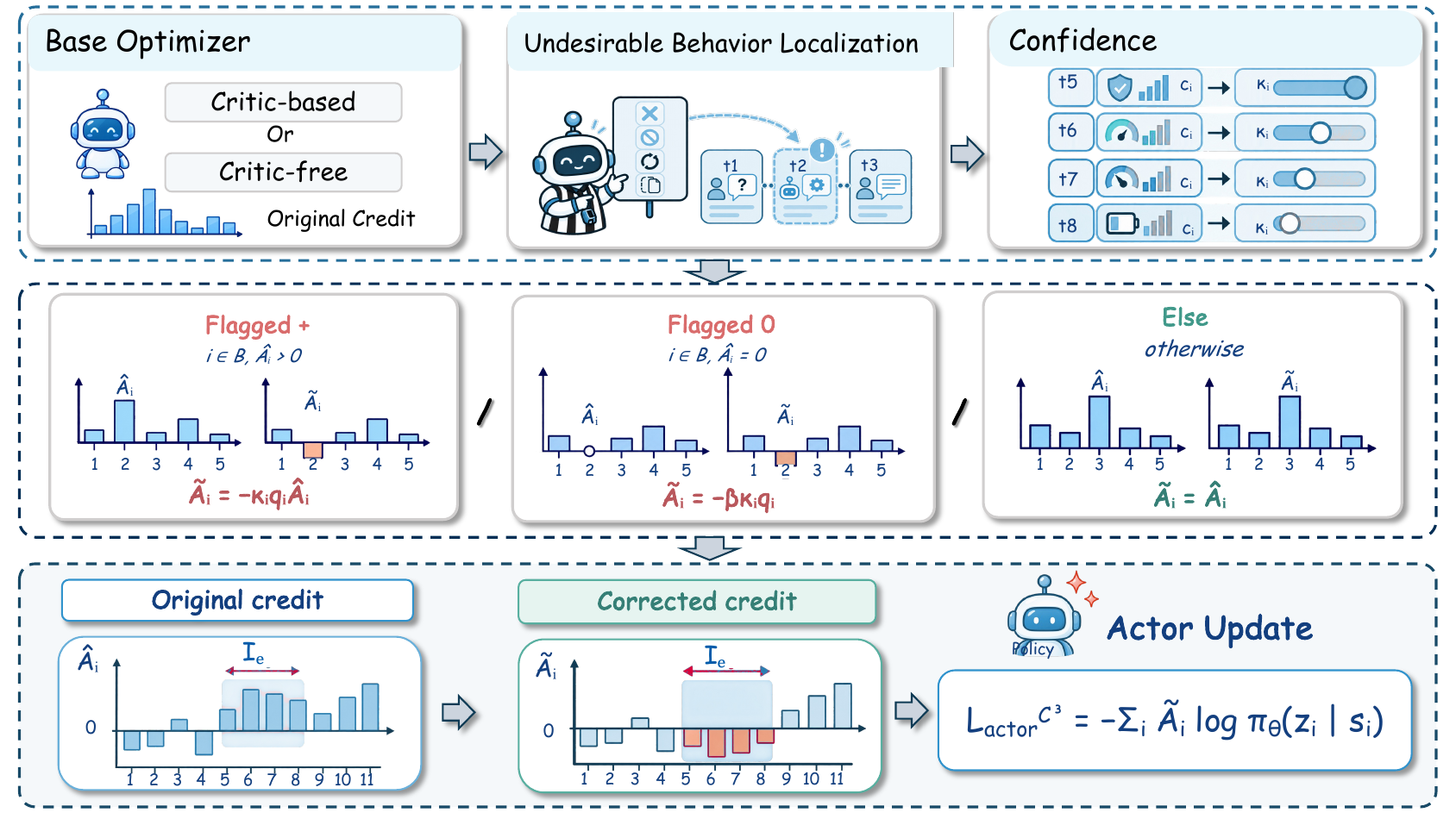}
	\caption{Overview of confidence-guided advantage correction for multi-turn agent reinforcement learning. A detector localizes an undesirable-behavior span, the actor-side correction is applied only within that span, and its strength follows the policy confidence.}
	\label{fig:method_overview}
\end{figure*}

We define \method{} as a post-estimation operator for multi-turn agent reinforcement learning. An external transcript detector localizes an \textit{undesirable behavior}, the base RL pipeline provides token-level advantages, and \method{} corrects these advantages before the actor update. We first motivate this design through an empirical behavioral and credit audit, and then define the inputs, localization rule, and correction operator. 
Figure~\ref{fig:method_overview} summarizes the resulting workflow.

\subsection{Motivation: Misguided Reinforcement and Local Credit}
\label{sec:pilot}

\begin{figure}[b]
	\centering
	\begin{subfigure}[t]{0.315\linewidth}
		\centering
		\includegraphics[width=\linewidth]{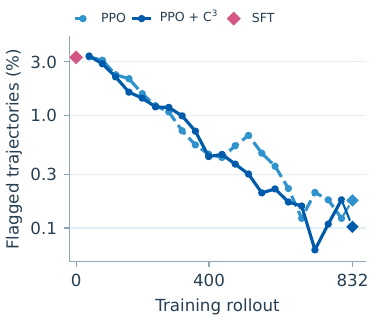}
		\caption{PPO training}
		\label{fig:pilot-ppo}
	\end{subfigure}
	\hfill
	\begin{subfigure}[t]{0.300\linewidth}
		\centering
		\includegraphics[width=\linewidth]{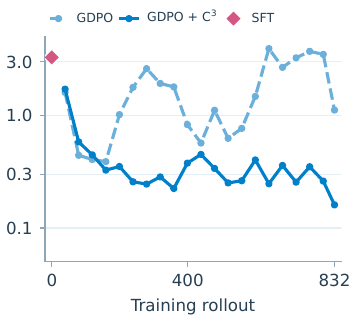}
		\caption{GDPO training}
		\label{fig:pilot-gdpo}
	\end{subfigure}
	\hfill
	\begin{subfigure}[t]{0.368\linewidth}
		\centering
		\includegraphics[width=\linewidth]{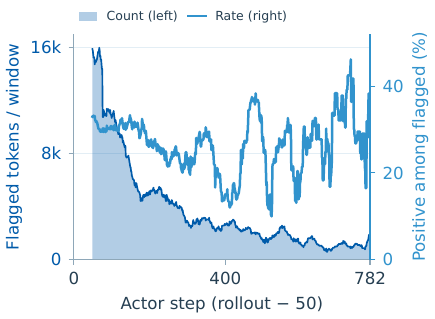}
		\caption{Positive-credit audit}
		\label{fig:pilot-credit}
	\end{subfigure}
	\caption{Misguided reinforcement during RL training.}
	\label{fig:pilot}
\end{figure}

We investigate local credit dynamics through two representative undesirable
behaviors, \textit{\textbf{invalid actions}} and \textit{\textbf{query loops}},
whose operational definitions and detection rules are provided in
Appendix~\ref{app:event_detection}. Figure~\ref{fig:intro}(a) illustrates
representative event-level examples, while Figure~\ref{fig:pilot} traces how
undesirable-behavior occurrence and token-level advantages evolve during RL
training.

\textbf{Observation 1: Undesirable behaviors reveal complex credit dynamics.}
As shown in Figure~\ref{fig:pilot}(a,b), undesirable-behavior occurrence
generally decreases during training for {critic-based PPO},
whereas {critic-free GDPO} shows no consistent decline and
even rebounds at the later stage. PPO derives token-level advantages through
a learned critic and GAE, while GDPO constructs relative advantages from
group-level trajectory signals that combine task outcomes with
undesirable-behavior penalties. These contrasting dynamics suggest that
the same undesirable behaviors can evolve differently under
different credit construction mechanisms, making local credit assignment
inherently nontrivial.

\textbf{Observation 2: Behavior suppression does not imply local credit alignment.}
Figure~\ref{fig:pilot}(c) shows that, among $63{,}511$ valid tokens within
detector-identified undesirable-behavior spans in an instrumented PPO run,
$18{,}460$ receive positive base advantage before correction, corresponding
to 29.07\%. Notably, this positive credit persists even as the number
of undesirable-behavior tokens decreases during training. This indicates that
reducing undesirable-behavior occurrence does not eliminate
incorrect local credit, suggesting that trajectory-level supervision can
still assign positive local credit to undesirable behaviors.

% \begin{tcolorbox}[
%     colback=gray!5,
%     colframe=black!70,
%     boxrule=0.6pt,
%     arc=2pt
% ]
% \textbf{Main Findings.}
% \textbf{(1) Local credit is mechanism-dependent:}
% the same undesirable behavior evolves differently under different credit
% construction mechanisms.
% \textbf{(2) Behavior reduction does not imply credit alignment:}
% undesirable behaviors can remain positively reinforced even as their
% occurrence declines.
% \end{tcolorbox}

% Together, these observations reveal a previously overlooked failure mode of
% multi-turn RL: \textit{\textbf{undesirable behaviors can continue to receive
% positive token-level advantages despite trajectory-level optimization}}. The
% contrasting PPO and GDPO dynamics further suggest that this failure mode is
% not tied to a single optimizer, but emerges from how local credit is
% constructed. 
% This motivates a minimal intervention: \textit{\textbf{instead
% of redesigning rewards or advantage estimation, can correcting positive
% credit on a small number of localized undesirable behaviors be sufficient to
% improve reinforcement learning?}}
{Together, these observations expose a failure mode in multi-turn RL: undesirable behaviors can remain positively reinforced at the token level even as they are suppressed at the trajectory level. The contrasting dynamics under PPO and GDPO further suggest that this issue is not specific to a particular credit construction mechanism, but can arise whenever trajectory-level signals are translated into local credit. This motivates a targeted intervention: }

\begin{tcolorbox}[
    colback=gray!5,
    colframe=black!70,
    boxrule=0.6pt,
    arc=2pt,
    left=6pt,
    right=6pt,
    top=5pt,
    bottom=5pt
]
\textit{Instead of redesigning rewards or advantage estimation, can correcting positive credit for a small number of localized undesirable behaviors be sufficient to improve RL?}
\end{tcolorbox}

\subsection{Local Credit Correction}
\label{sec:correction}
{Motivated by the local credit misalignment identified above, we aim to correct policy credit only where its direction is inconsistent with the detected local behavior, while preserving the credit assigned to the remaining parts of the trajectory. To this end, C$^3$ operates on the token-level advantages $\hat{A}_i$ produced by the base RL optimizer and the localized failure-token set $\mathcal{B}$.}

{Formally, for a detected token $i\in\mathcal{B}$ with $\hat{A}_i>0$, we reverse the direction of its advantage and scale the correction by a token-specific gain:
\begin{equation}
    \tilde{A}_i
    =
    -\alpha_i \hat{A}_i,
    \qquad
    i\in\mathcal{B},\ \hat{A}_i>0,
    \label{eq:basic_correction}
\end{equation} 
where $\alpha_i$ controls the strength of the correction. Tokens whose advantages are already negative require no directional correction and are left unchanged.}

\paragraph{\textbf{Confidence-Guided Correction Strength.}}
{Localization determines \emph{where} credit should be corrected, while the correction gain determines \emph{how strongly} the intervention should be applied. We parameterize this gain using both the severity of the detected failure and the behavior policy's confidence in the corresponding action.}

{For each detected token $i$, let $J_{e(i)}$ denote the action span associated with the failure event covering that token. We measure the confidence of the behavior-policy snapshot $\pi_{\theta_k}$ using one of three aggregation strategies:
\begin{equation}
	\left\{
	\begin{aligned}
		c_i^{\mathrm{token}}(\theta_k)
		&= \pi_{\theta_k}(z_i\mid s_i),
		&& \\
		c_i^{\mathrm{arith}}(\theta_k)
		&= \frac{1}{|J_{e(i)}|}
		\sum_{j\in J_{e(i)}}\pi_{\theta_k}(z_j\mid s_j),
		&& \\
		c_i^{\mathrm{geo}}(\theta_k)
		&= \exp\left(
		\frac{1}{|J_{e(i)}|}
		\sum_{j\in J_{e(i)}}
		\log \pi_{\theta_k}(z_j\mid s_j)
		\right)
		&& .
	\end{aligned}
	\right.
	\label{eq:confidence}
\end{equation}}

{We use the arithmetic span mean as the default confidence measure. For span-level aggregation, all tokens associated with the same failure event share the same confidence value, whereas token probability provides token-specific confidence. Since $\pi_{\theta_k}$ is the behavior-policy snapshot used to generate the trajectory, the confidence adapts as the policy evolves during training. During actor optimization, the confidence is detached from the computation graph and acts only as an adaptive correction coefficient.}

{For the main method, the correction gain is the selected policy confidence:
\begin{equation}
    \kappa_i
    =
    c_i(\theta_k).
    \label{eq:kappa}
\end{equation}
Combining policy confidence with the event severity $q_i$, the final correction strength is
\begin{equation}
    \alpha_i = \kappa_i q_i .
    \label{eq:correction_strength}
\end{equation}}

\paragraph{\textbf{Complete Correction Operator.}}
{The resulting corrected advantage is
\begin{equation}
    \tilde{A}_i =
    \begin{cases}
        -\kappa_i q_i \hat{A}_i,
        & i\in\mathcal{B},\ \hat{A}_i>0,\\[3pt]
        -\beta\kappa_i q_i,
        & i\in\mathcal{B},\ \hat{A}_i=0,\\[3pt]
        \hat{A}_i,
        & \text{otherwise}.
    \end{cases}
    \label{eq:c3}
\end{equation}
Thus, C$^3$ selectively reverses positive credit assigned to detected failure behaviors, while preserving credit that is already negative or unrelated to the detected failure. The optional coefficient $\beta\geq0$ handles exact zero-advantage cases. We set $\beta=0$ in all main experiments, such that zero-advantage tokens are also left unchanged.}

\subsection{Integration with Policy Optimization}
\label{sec:integration}

{Building on the simple advantage correction rule above, we next integrate it into the RL optimization process. By applying the correction directly to the actor-side learning signal, C$^3$ preserves the original reward and credit-construction procedure while allowing the same mechanism to be used across different optimization paradigms. We next describe how this correction affects the actor update and how it is instantiated in critic-based and critic-free RL.}
\paragraph{\textbf{Effect on the Actor Update.}}
{Ignoring clipping and regularization terms, the actor objective using the
corrected advantages is
\begin{equation}
    \mathcal{L}_{\mathrm{actor}}^{\mathrm{C^3}}
    =
    -\sum_i
    \tilde{A}_i
    \log\pi_\theta(z_i\mid s_i).
    \label{eq:actor_loss}
\end{equation}
For a detected token with $\hat{A}_i>0$, C$^3$ changes its contribution
from
\begin{equation}
    -\hat{A}_i\log\pi_\theta(z_i\mid s_i)
\end{equation}
to
\begin{equation}
    \kappa_iq_i\hat{A}_i
    \log\pi_\theta(z_i\mid s_i),
\end{equation}
thereby reversing the direction of its policy-gradient contribution. Tokens outside the detected failure spans, as well as detected tokens with negative advantages, retain their original actor-side credit.}

\paragraph{\textbf{Critic-Based and Critic-Free Integration.}}
{For critic-based optimization, C$^3$ is applied after the original advantage estimation and actor-side normalization. In our PPO implementation, this corresponds to applying Eq.~\ref{eq:c3} after advantage whitening. The critic continues to be trained using the original return targets, so C$^3$ affects only the actor update.}

{For critic-free group-relative optimization, C$^3$ is applied after group-relative advantage construction, leaving the underlying credit construction unchanged. It corrects only the resulting actor-side advantages before the policy update, making the same operator applicable across different credit-construction paradigms.}

\begin{algorithm}[t]
\caption{RL Optimization with Confidence-Guided Credit Correction (C$^3$)}
\label{alg:c3_rl}
\begin{algorithmic}[1]
\REQUIRE Actor policy $\pi_\theta$, base RL optimizer $\mathcal{O}$, detector $\mathcal{D}$
\ENSURE Updated policy $\pi_\theta$

\STATE Roll out trajectories under $\pi_{\theta_k}$ and compute base advantages $\{\hat A_i\}$ using $\mathcal{O}$
\STATE Detect events $\mathcal{E}\leftarrow\mathcal{D}(\tau)$ and localize $(I_e,J_e,q_e)$
\STATE Initialize corrected advantages $A_i\leftarrow\hat A_i$ for all tokens

\FOR{each event $e\in\mathcal{E}$}
    \STATE $\ell_j=\log\pi_{\theta_k}(z_j\mid s_j)$ for $j\in J_e$
    \STATE $c_e\leftarrow\frac{1}{|J_e|}\sum_{j\in J_e}\exp(\ell_j)$
    \FOR{each token $i\in I_e$}
        \STATE $q_i\leftarrow\max(q_i,q_e)$
        \IF{$\hat A_i>0$}
            \STATE $A_i\leftarrow-q_i c_e\hat A_i$
        \ENDIF
    \ENDFOR
\ENDFOR

\STATE Update $\pi_\theta$ using corrected advantages $\{A_i\}$
\end{algorithmic}
\end{algorithm}
\section{Experiments}
\label{sec:experiments}
%We evaluate C$^3$ from three complementary perspectives. We first examine whether C$^3$ consistently improves agent performance under different policy optimization paradigms.
%We then investigate how C$^3$ changes undesirable behaviors and their local credit during training to understand its effect on the underlying optimization process. 
%Finally, we evaluate C$^3$ on two multi-turn agent tasks, search question answering and WebShop, to examine its effectiveness across different task environments.

\subsection{Experimental Setup}
\label{subsec:setup}

\noindent \textbf{Implementation Details.}
For multi-turn search question answering, we use Qwen3-8B~\citep{qwen3} with the SLIME framework~\citep{slime}. The agent interacts with a retrieval environment built on the Wiki-Fixed corpus introduced in~\citep{retrieve}, using Qwen3-Embedding-8B~\citep{qwen3_embedding} with 4096-dimensional embeddings and FAISS for retrieval. For WebShop~\citep{webshop}, we use Qwen3-4B with full-parameter updates in the standard WebShop environment. All models are trained under their corresponding task protocols. 
%Further implementation details are provided in Appendix~\ref{app:reward_config}.

\noindent \textbf{Baselines.}
We evaluate \method{} under two optimization families: \textit{\textbf{critic-based optimization}} and \textit{\textbf{critic-free group-relative optimization}}. For critic-based optimization, we compare standard PPO, PPO with a turn-level process reward, and PPO with \method{}. The process-reward baseline incorporates detector events into return and advantage estimation, whereas \method{} directly corrects token-level advantages on detected spans. For critic-free group-relative optimization, we compare GDPO~\citep{gdpo} with and without \method{}. GDPO follows the group-relative paradigm of GRPO but decouples the normalization of individual reward components. Configurations are provided in Appendix~\ref{app:reward_credit}.

\noindent \textbf{Evaluation Benchmarks and Metrics.}
For search question answering, we evaluate on seven multi-turn benchmarks: NQ~\citep{nq}, TriviaQA~\citep{triviaqa}, PopQA~\citep{popqa}, HotpotQA~\citep{hotpotqa}, 2WikiMultiHopQA~\citep{twowiki}, MuSiQue~\citep{musique}, and Bamboogle~\citep{bamboogle}. We report strict exact match (EM), with the overall score computed as the unweighted mean across the seven benchmarks. 
For WebShop, we report the task score and success rate on the fixed
500-instance benchmark split.

\subsection{Main Results}
\label{subsec:main_results}

We evaluate C$^3$ on two multi-turn agent environments, Search QA and WebShop, under both critic-based PPO and critic-free GDPO, reporting both final task performance and training dynamics.

\noindent\textbf{Search Question Answering.}
Table~\ref{tab:main_search} reports strict EM on seven multi-turn question-answering benchmarks. 
The complete benchmark-level evaluation dynamics are provided in Appendix~\ref{subsec:seven_benchmark_dynamics}.
Under PPO, C$^3$ improves the average score from $47.79\%$ to \textbf{$50.20\%$}, yielding an absolute gain of \textbf{2.41 points}, with improvements on six of the seven benchmarks. The largest gain is observed on 2Wiki, where C$^3$ improves strict EM by 11.5 points, followed by a 2.6-point gain on Bamboogle. Under GDPO, C$^3$ improves the average score from $48.37\%$ to \textbf{$49.23\%$}, an absolute gain of \textbf{0.86 points}, with improvements on six of the seven benchmarks.

\begin{table*}[htbp]
	\centering
	\footnotesize
\caption{Strict EM (\%) on seven QA benchmarks with
		Qwen3-8B. All entries report mean $\pm$ standard deviation over four
		independent samples at temperature $0.8$ with top-$p=0.95$.
		\textbf{Bold} marks the best mean within each optimization family.}
	\label{tab:main_search}
	\setlength{\tabcolsep}{2.5pt}
	\renewcommand{\arraystretch}{1.15}
	\begin{tabular}{l c c c c c c c c}
		\toprule
		\multirow{2}{*}{Method} &
		\multicolumn{3}{c}{Single-Hop QA} &
		\multicolumn{4}{c}{Multi-Hop QA} &
		\multirow{2}{*}{Avg.} \\
		\cmidrule(lr){2-4} \cmidrule(lr){5-8}
		& NQ & TriviaQA & PopQA & HotpotQA & 2Wiki & Musique & Bamb. & \\
		\midrule
	SFT
		& \meanstd{35.2}{0.2} & \meanstd{59.2}{0.2} & \meanstd{37.8}{0.0}
		& \meanstd{48.9}{0.2} & \meanstd{64.1}{0.3} & \meanstd{22.6}{0.2}
		& \meanstd{46.2}{1.2} & \meanstd{44.87}{0.17} \\
		\midrule
		
		\rowcolor{PPOFamily}
		\multicolumn{9}{l}{\textit{~~~Critic-Based Optimization}} \\
	PPO 
	& \meanstd{40.7}{0.5} & \meanstd{62.1}{0.2} & \meanstd{43.4}{0.1}
	& \meanstd{55.2}{0.2} & \meanstd{56.4}{0.3} & \meanstd{28.8}{0.5}
	& \meanstd{47.8}{2.8}
	& \meanstd{47.79}{0.50} \\
	\quad + process reward
	& \textbf{42.0}\scriptsize$\pm$0.4 & \meanstd{62.3}{0.3} & \meanstd{43.5}{0.2}
	& \meanstd{55.0}{0.1} & \meanstd{60.8}{0.3} & \meanstd{23.9}{0.6}
	& \meanstd{42.2}{2.4}
	& \meanstd{47.10}{0.43} \\
	\quad + \method{}
	& \meanstd{39.5}{0.3} & \textbf{63.5}\scriptsize$\pm$0.1 & \textbf{45.0}\scriptsize$\pm$0.3
	& \textbf{55.6}\scriptsize$\pm$0.1 & \textbf{67.9}\scriptsize$\pm$0.3
	& \textbf{29.7}\scriptsize$\pm$0.1 & \textbf{50.4}\scriptsize$\pm$0.0
	& \textbf{50.20}\scriptsize$\pm$0.10 \\
		\midrule
		
		\rowcolor{GDPOFamily}
		\multicolumn{9}{l}{\textit{~~~Group-Relative Optimization}} \\
	GDPO
	& \meanstd{40.0}{0.4} & \meanstd{61.0}{0.2} & \meanstd{43.2}{0.2}
	& \meanstd{55.5}{0.1} & \textbf{63.1}\scriptsize$\pm$0.2
	& \meanstd{29.1}{0.5} & \meanstd{46.8}{1.4}
	& \meanstd{48.37}{0.19} \\
	\quad + \method{}
	& \textbf{41.6}\scriptsize$\pm$0.3 & \textbf{62.8}\scriptsize$\pm$0.2
	& \textbf{45.7}\scriptsize$\pm$0.3 & \textbf{55.7}\scriptsize$\pm$0.3
	& \meanstd{60.9}{0.1} & \textbf{29.5}\scriptsize$\pm$0.6
	& \textbf{48.4}\scriptsize$\pm$2.5
	& \textbf{49.23}\scriptsize$\pm$0.35 \\
		\bottomrule
	\end{tabular}
\end{table*}
% \vspace{-5pt}

\begin{figure*}[t]
	\centering
	\begin{subfigure}[t]{0.245\linewidth}
		\centering
		\includegraphics[width=\linewidth]{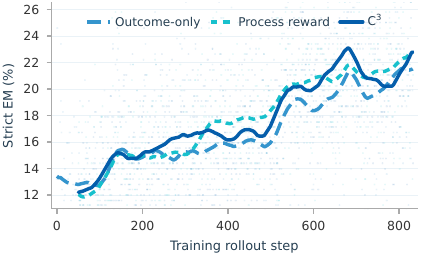}
		\caption{PPO training}
		\label{fig:em-ppo-train}
	\end{subfigure}\hfill
	\begin{subfigure}[t]{0.245\linewidth}
		\centering
		\includegraphics[width=\linewidth]{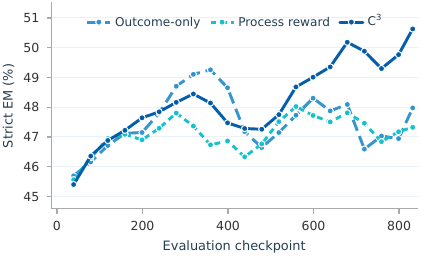}
		\caption{PPO evaluation}
		\label{fig:em-ppo-eval}
	\end{subfigure}\hfill
	\begin{subfigure}[t]{0.245\linewidth}
		\centering
		\includegraphics[width=\linewidth]{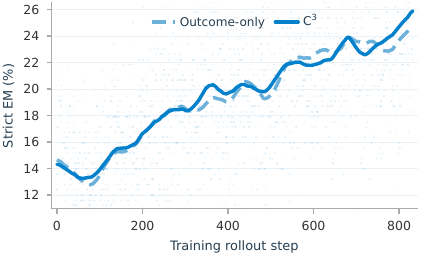}
		\caption{GDPO training}
		\label{fig:em-gdpo-train}
	\end{subfigure}\hfill
	\begin{subfigure}[t]{0.245\linewidth}
		\centering
		\includegraphics[width=\linewidth]{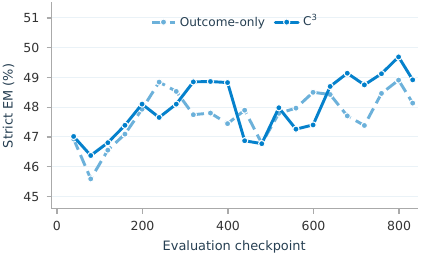}
		\caption{GDPO evaluation}
		\label{fig:em-gdpo-eval}
	\end{subfigure}

	\caption{Panels (a,c) show smoothed training EM, and panels (b,d) show evaluation strict EM.}
	\label{fig:search_em_dynamics}
\end{figure*}

 Figure~\ref{fig:search_em_dynamics} shows the corresponding training and evaluation dynamics across the seven benchmarks.
During training, C$^3$ generally reaches higher benchmark-level performance than the corresponding baselines, although the magnitude and timing of the gains vary across tasks. The periodic evaluation curves show a similar pattern at the aggregate level, with \textbf{\textit{C$^3$ achieving higher evaluation performance at multiple stages of training}} rather than only at the final checkpoint.

\begin{figure}[H]
  \centering
  \begin{minipage}[t]{0.3\linewidth}
	\centering
	\footnotesize
	\vspace{0pt}
\captionof{table}{Best benchmark performance.}
	\label{tab:webshop}
	\setlength{\tabcolsep}{2pt}
	\renewcommand{\arraystretch}{1.12}
%	\begin{tabular*}{\linewidth}{@{\extracolsep{\fill}}lcc@{}}
%	\toprule
	%Method & Score & Success \\
	%\midrule
	%SFT  & 67.77 & 59.2 \\
	%\midrule
	%\rowcolor{WebShopBlue}
	%\multicolumn{3}{l}{\textit{Critic-Based Optimization}} \\
	%	PPO & 80.59 & 67.4 \\
		%\quad + process & 77.81 & 69.0 \\
		%\quad + \method{} & \textbf{83.71} & \textbf{70.8} \\
		%\midrule
		%\rowcolor{WebShopBlue}
%	\multicolumn{3}{l}{\textit{Group-Relative Optimization}} \\
		%GDPO & 84.87 & 68.8 \\
	%	\quad + \method{} & \textbf{86.29} & \textbf{71.4} \\
	%	\bottomrule
%	\end{tabular*}
\begin{tabular*}{\linewidth}{@{\extracolsep{\fill}}lcc@{}}
	\toprule
	Method & Score & Success \\
	\midrule
	
	\cellcolor{WebShopGray!20} SFT
	& 67.77
	& 59.2 \\
	\midrule
	
	\cellcolor{WebShopBlue} PPO
	& 80.59
	& 67.4 \\
	\quad + process & 77.81 & 69.0 \\
	\quad + \method{} & \textbf{83.71} & \textbf{70.8} \\
	\midrule
	
	\cellcolor{WebShopBlue} GDPO
	& 84.87
	& 68.8 \\
	\quad + \method{} & \textbf{86.29} & \textbf{71.4} \\
	\bottomrule
\end{tabular*}
\end{minipage}
  \hfill
  \begin{minipage}[t]{0.69\linewidth}
    \vspace{0pt}
    \centering
    \setcounter{subfigure}{0}
    \begin{subfigure}[t]{0.49\linewidth}
      \centering
      \includegraphics[width=\linewidth]{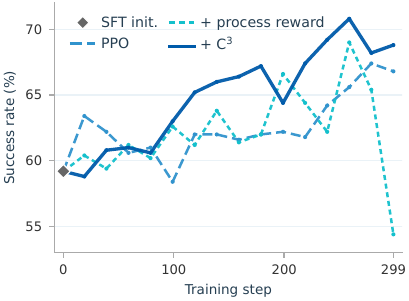}
      %\caption{PPO}
    \end{subfigure}
    \hfill
    \begin{subfigure}[t]{0.49\linewidth}
      \centering
      \includegraphics[width=\linewidth]{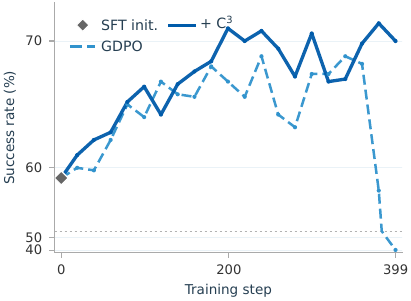}
      %\caption{GDPO}
    \end{subfigure}
    % \vspace{-6pt}
    % \captionsetup{skip=2pt}
   \captionof{figure}{WebShop validation trajectories during training.}
    \label{fig:webshop_trajectory}
  \end{minipage}
\end{figure}

% \vspace{-5pt}
\noindent\textbf{WebShop.}
We further evaluate C$^3$ on WebShop, a second multi-turn agent environment with a different task structure. Table~\ref{tab:webshop} includes the common SFT initialization at step 0 and reports task score and success rate at each trained run's best-success checkpoint. Under PPO, C$^3$ improves the peak success rate from $67.4\%$ to \textbf{$70.8\%$}, with the task score increasing from $80.59$ to \textbf{$83.71$}. 

Under GDPO, C$^3$ improves success from $68.8\%$ to \textbf{$71.4\%$} and the task score from $84.87$ to \textbf{$86.29$}. Figure~\ref{fig:webshop_trajectory} shows the corresponding training trajectories. Notably, PPO with process reward and GDPO both exhibit substantial performance degradation in the later stages of training, whereas neither PPO+C$^3$ nor GDPO+C$^3$ shows a comparable collapse within the evaluated training windows.

Taken together, these results show that C$^3$ improves performance under both critic-based and critic-free optimization across the two multi-turn agent environments.
The gains are reflected not only in final benchmark performance, but also in the training dynamics, where C$^3$ \textbf{\textit{maintains stronger performance for longer before the late-stage degradation observed in the corresponding baselines}}.

\section{Analysis}
\label{sec:analysis}
We further analyze how C$^3$ changes agent behavior and evaluate its key
design choices through undesirable-behavior analysis and confidence
aggregation ablations. Additional experiments and case studies are provided
in Appendices~\ref{app:additional_experiments} and
\ref{app:case_studies}.

% We further analyze how C$^3$ affects agent behavior across optimization paradigms and ablate confidence aggregation design. Additional experiments and cases are provided in Appendices~\ref{app:additional_experiments} and~\ref{app:case_studies}.

\subsection{Does C\texorpdfstring{$^3$}{3} Change Agent Behavior?}
\label{subsec:analysis_badpatterns}

\begin{figure}[htbp]
	\centering
	\begin{subfigure}[t]{0.315\linewidth}
		\centering
		\includegraphics[width=\linewidth]{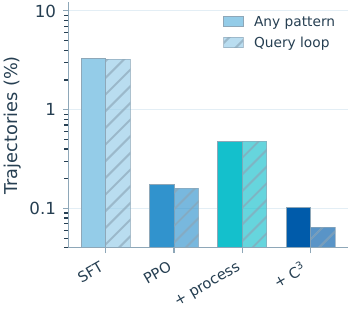}
		\caption{Undesirable-behavior rate}
		\label{fig:eval-badpatterns-rate}
	\end{subfigure}
	\hfill
	\begin{subfigure}[t]{0.315\linewidth}
		\centering
		\includegraphics[width=\linewidth]{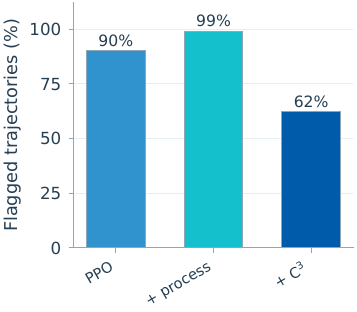}
		\caption{Query-loop share}
		\label{fig:eval-badpatterns-share}
	\end{subfigure}
	\hfill
	\begin{subfigure}[t]{0.315\linewidth}
		\centering
		\includegraphics[width=\linewidth]{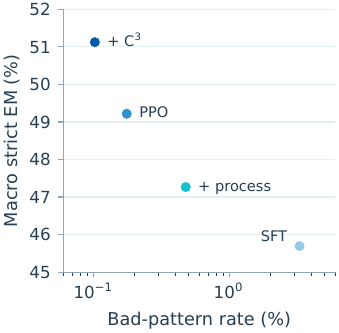}
		\caption{Rate vs.\ accuracy}
		\label{fig:eval-badpatterns-accuracy}
	\end{subfigure}
	\caption{Greedy evaluation on $51{,}713$ questions per arm.}
	\label{fig:eval_badpatterns}
\end{figure}

We examine SFT, PPO, PPO with process reward, and PPO+\method{} on the same
$51{,}713$ Search QA questions under greedy decoding. Figure~\ref{fig:eval_badpatterns}
separates invalid actions from query loops to examine how different types of
undesirable behaviors respond to outcome-level training.

PPO substantially reduces invalid actions relative to SFT, suggesting that
straightforward tool-call failures can be effectively suppressed by
trajectory-level training. Query loops, however, show a different pattern:
the process-reward variant reaches a query-loop rate of $0.474\%$, nearly
$3\times$ that of PPO ($0.159\%$), and 245 of its 247 flagged trajectories
contain query loops. In contrast, PPO+\method{} reduces the query-loop rate
to $0.064\%$, a $2.5\times$ reduction relative to PPO and a $7.4\times$
reduction relative to the process-reward variant.
These results suggest that \textit{\textbf{trajectory-level training is
effective for explicit action failures but less effective for redundant
cross-turn behavior}}. C$^3$ complements this by correcting positive local
credit on detected behavior spans, providing a more direct signal for
behaviors that are not reliably suppressed by trajectory-level supervision.

\subsection{Ablation of C\texorpdfstring{$^3$}{3} Design Choices}
\label{subsec:analysis_confidence}

\begin{figure}[htbp]
	\centering
	\begin{minipage}[t]{0.42\linewidth}
		\vspace{0pt}
		\centering
		\captionof{table}{Final-checkpoint macro strict EM (\%) for PPO confidence
			aggregation.}
		\label{tab:confidence_ablation}
		\footnotesize
		\setlength{\tabcolsep}{2pt}
		\renewcommand{\arraystretch}{1.12}
		\begin{tabular*}{\linewidth}{@{\extracolsep{\fill}}lc@{}}
			\toprule
			Confidence rule & Macro strict EM (\%) \\
			\midrule
			Outcome-only reference & 47.61 \\
			Token probability & 45.21 \\
			Arithmetic span mean & \textbf{50.57} \\
			Geometric span mean & 48.94 \\
			\bottomrule
		\end{tabular*}
	\end{minipage}
	\hfill
	\begin{minipage}[t]{0.57\linewidth}
		\vspace{0pt}
		\centering
		\includegraphics[width=\linewidth]{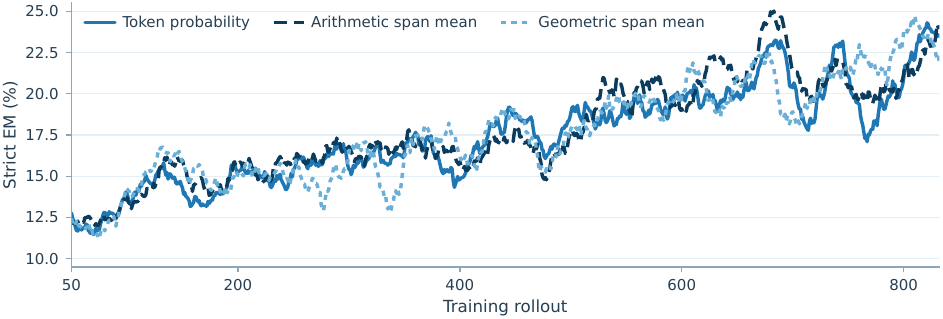}
		\captionof{figure}{Smoothed training strict EM for
			different confidence aggregation strategies under PPO.}
		\label{fig:confidence_modes_training}
	\end{minipage}
\end{figure}

C$^3$ uses policy confidence to determine the strength of local credit
correction. We compare token-level confidence with two span-level aggregation
strategies. Table~\ref{tab:confidence_ablation} reports the final macro strict
EM, and Figure~\ref{fig:confidence_modes_training} shows the training
trajectories.

Arithmetic span averaging achieves the highest final macro strict EM at
\textbf{$50.57\%$}, outperforming geometric span averaging ($48.94\%$),
token-level probability ($45.21\%$), and the outcome-only reference
($47.61\%$), indicating that \textit{\textbf{aggregating confidence over the
entire undesirable-behavior span provides a more effective correction signal
than token-level confidence}}. Figure~\ref{fig:confidence_modes_training}
shows the same pattern throughout training: token-level confidence
consistently underperforms both span-level variants, whereas arithmetic and
geometric averaging remain nearly overlapping. Since token probability often
reflects local generation uncertainty rather than event reliability,
token-level correction can overweight formatting tokens, transition words,
punctuation, or closing tags within a detected span. Thus,
\textit{\textbf{the key design choice is span-level rather than token-level
confidence aggregation}}, while the difference between arithmetic and
geometric averaging remains minor.
% \section{Conclusion}

% Outcome-only supervision can assign positive advantages to spans corresponding
% to undesirable local behaviors, creating a mismatch between behavioral quality
% and policy credit. \method{} addresses this issue through localized credit
% correction: it reverses positive advantages on detected undesirable-behavior
% spans and scales the correction by detached behavior-policy confidence, while
% leaving the task reward, returns, and critic targets unchanged. Across the
% reported PPO and group-relative experiments, \method{} improves average exact
% match and reduces detected undesirable behaviors, while the comparison with
% turn-level process reward highlights the importance of how process supervision
% is incorporated into credit construction. Overall, these results suggest that
% {{effective process supervision is not only about providing
% 		denser behavioral feedback, but also about translating that feedback into
% 		appropriate local policy credit}}.

\section{Conclusion}

Trajectory-level supervision can assign positive advantages to spans
corresponding to undesirable local behaviors, creating a mismatch between
behavioral quality and policy credit. \method{} addresses this through
confidence-guided local credit correction, reversing positive advantages on
detected undesirable-behavior spans while leaving the task reward, returns,
and critic targets unchanged. Across experiments,
\method{} improves average exact match and reduces detected undesirable
behaviors, highlighting that {effective process supervision is
not only about providing denser behavioral feedback, but also about
translating that feedback into appropriate local policy credit}.

% \section*{Limitations}

% \noindent\textbf{Behavior Coverage.}
% Our experiments focus on two types of undesirable behaviors, invalid actions
% and query loops, in multi-turn tool-use tasks. These behaviors cover both
% action-level failures and redundant cross-turn interactions, which are the
% primary behavioral patterns examined in our study. Extending the same
% credit-correction framework to other types of undesirable behaviors remains a
% natural direction for future work.

% \noindent\textbf{Confidence Aggregation.}
% C$^3$ determines the correction strength from policy confidence, and our
% experiments consider token-level probability, arithmetic span-level
% averaging, and geometric span-level averaging. The span-level variants share
% the same localized behavior span while differing in how confidence is
% aggregated across its tokens. Other aggregation choices can be incorporated
% without changing the underlying credit-correction mechanism.

% \noindent\textbf{Model and Benchmark Scope.}
% Due to computational resource constraints, our experiments use Qwen3-4B and
% Qwen3-8B across the Search QA and WebShop benchmarks. The two model sizes
% cover both the search-oriented and interactive settings considered in this
% work, while keeping the training configurations consistent across the
% corresponding experimental arms. Evaluations at additional model scales and
% on other agent benchmarks remain outside the scope of the current study.

\subsubsection*{AI Use Statement}

During this work, we used LLM for dataset construction to generate
candidate SFT trajectories for Search QA and WebShop under predefined
prompting protocols. All generated trajectories were subsequently filtered
using task-specific quality constraints before being included in the final
training corpus. During manuscript preparation, we also used an LLM-based
assistant for language editing, LaTeX troubleshooting, and local organization
of supplementary material. All technical content, experimental design,
analysis, and final manuscript revisions were reviewed and verified by the
authors.

\subsubsection*{Ethics Statement}

\method{} is designed to improve the reliability of tool-integrated LLM
agents by discouraging invalid, unsupported, and redundant actions. Since
these agents may interact with external environments, deployment should
retain appropriate sandboxing, tool-permission controls, audit logs, and
task-specific safety checks. Our experiments use agent-generated trajectory
transcripts and do not involve human subject data. Any automated assistance
used in preparing this manuscript was reviewed by the authors for factual
accuracy and compliance with venue policies.

\subsubsection*{Reproducibility Statement}

\method{} is implemented as a post-estimation operation on token-level
advantages and does not require changes to the reward function, rollout
collection, or critic. It can therefore be integrated into a policy-gradient
implementation by applying the correction after advantage normalization and
before the actor update. Across experiments, the compared arms share the same
task data, SFT initialization within each setting, terminal task reward,
prompt protocol, turn budget, detectors, and evaluation procedure. An
anonymous code repository containing the implementation and training
configuration is available at
\url{https://anonymous.4open.science/r/C3-CC78}; the full repository will be
released with the camera-ready version.

\bibliography{custom}
\bibliographystyle{iclr2027_conference}

\appendix
\clearpage

% The appendix is organized as data, implementation, additional experiments,
% and case studies so that reproduction details are separated from examples.
\section{Data Construction}
\label{app:data_construction}

\subsection{Search QA: 7-BMK}
\label{app:data_search}

\textbf{Evaluation Benchmark.}
We evaluate Search QA on seven multi-turn question-answering benchmarks:
NQ~\citep{nq}, TriviaQA~\citep{triviaqa}, PopQA, HotpotQA~\citep{hotpotqa},
2WikiMultiHopQA~\citep{twowiki}, MuSiQue~\citep{musique}, and
Bamboogle~\citep{bamboogle}. The fixed evaluation suite contains 51,713
questions in total. We use the Wiki-Fixed corpus~\citep{retrieve} for passage
retrieval and report strict exact match (EM) for each benchmark, with the
aggregate score computed as the unweighted mean across the seven benchmarks.
All compared methods are evaluated on the same benchmark suite under the same
evaluation protocol.

\subsection{WebShop}
\label{app:data_webshop}

\textbf{Evaluation Benchmark.}
We use WebShop~\citep{webshop} as a second multi-turn environment. Following
the official WebShop protocol, training and evaluation use disjoint task
splits. All reported main results are evaluated on the fixed 500-task
benchmark, with task IDs held constant across all runs. A task is considered
successful only when the environment accepts the final purchase; we
additionally report the environment-provided task score.

\subsection{SFT Initialization and RL Rollouts}
\label{app:data_training}

\begin{table}[H]
	\centering
	\small
	\caption{Summary of data scales used for SFT, RL training, and evaluation.}
	\label{tab:data_construction}
	\setlength{\tabcolsep}{5pt}
	\renewcommand{\arraystretch}{1.15}
	\begin{tabular}{@{}llr@{}}
		\toprule
		Stage & Dataset & Size \\
		\midrule
		\multirow{7}{*}{Evaluation}
		& NQ & 3,610 \\
		& TriviaQA & 11,313 \\
		& PopQA & 14,267 \\
		& HotpotQA & 7,405 \\
		& 2WikiMultiHopQA & 12,576 \\
		& MuSiQue & 2,417 \\
		& Bamboogle & 125 \\
		\cmidrule(lr){2-3}
		& \textbf{7-BMK Total} & \textbf{51,713} \\
		& WebShop Benchmark & \textbf{500} \\
		\midrule
		\multirow{2}{*}{SFT}
		& Search QA & 3,000 $\rightarrow$ 9,329 rows \\
		& WebShop & 200 $\rightarrow$ 1,713 rows \\
		\midrule
		\multirow{2}{*}{RL}
		& Search QA & 66,309 $\rightarrow$ 265,236 $\rightarrow$ $\approx$50,157 \\
		& WebShop & Official training split \\
		\bottomrule
	\end{tabular}
\end{table}

\textbf{Search QA SFT.}
We construct the Search QA cold-start SFT corpus using GLM-5.2
\citep{glm5team2026}. Starting from 3,000 deduplicated source questions,
GLM-5.2 generates executable search-and-answer traces under a strict-multihop
contract, emphasizing evidence decomposition, one action per turn, and answer
granularity. We retain traces that are strict-exact-match, format-valid,
search-first, non-repeating, and free of invalid actions. The retained
trajectories are deduplicated by source trajectory and normalized question,
then converted into 9,329 action-level continuation rows, corresponding to an
average of 3.11 supervised actions per source question.

\textbf{WebShop SFT.}
We construct a separate WebShop append-SFT corpus using GLM-5.2
\citep{glm5team2026}. It contains 200 unique strict-success trajectories. We
retain only trajectories for which the environment accepts the purchase and
the trace contains no invalid or non-clickable action, server error, or
missing task key. Each retained trajectory is expanded into one action-level
row with the complete preceding history, current observation, admissible
actions, and teacher action as the target, yielding 1,713 training rows.

\textbf{Search QA RL Rollouts.}
The Search QA RL pool begins with 66,309 prompt-fixed residual questions.
Four responses are sampled for each prompt, yielding 265,236 sampled
responses. After filtering all-correct groups, approximately 50,157
prompt-response records remain for RL training. Each record contains the
prompt, response, task reward, and process metadata. The rollout prompt set,
task IDs, and evaluation benchmark are held fixed across optimizer arms.

\textbf{RL Initialization.}
Search QA RL runs start from the Qwen3-8B Search QA SFT checkpoint, while
WebShop RL runs start from the Qwen3-4B WebShop append-SFT checkpoint. These
SFT checkpoints provide the common initialization for the corresponding RL
runs.

Table~\ref{tab:data_construction} summarizes the data scales used for SFT, RL
training, and evaluation.

\section{Implementation Details}
\label{app:implementation}

\subsection{Event Detection}
\label{app:event_detection}

\textbf{Action Parsing.}
The search agent emits one action per turn. A valid turn contains either a
search action,
\[
\begin{aligned}
	&\texttt{<think>...</think>}\\[-1pt]
	&\qquad\texttt{<search>...</search>},
\end{aligned}
\]
or a final answer action,
\[
\begin{aligned}
	&\texttt{<think>...</think>}\\[-1pt]
	&\qquad\texttt{<answer>...</answer>}.
\end{aligned}
\]
No text is allowed outside the corresponding tags. The parser performs strict
validation together with a single-action recovery pass, so that recoverable
formatting errors remain executable actions. Errors that affect only
strict-format evaluation are retained in the evaluation logs but are not
treated as undesirable-behavior events.

\textbf{Behavior Categories.}
We consider two types of undesirable behaviors: \textit{\textbf{invalid
		actions}} and \textit{\textbf{query loops}}. Invalid actions include malformed
or impermissible tool interactions, such as unclosed tool-call tags,
unavailable tools, invalid tool arguments, or actions issued after a
completed answer. Query loops refer to redundant tool interactions in which
the same or nearly identical query is repeated across turns. The first
occurrence is retained, while each subsequent occurrence is treated as an
undesirable-behavior event.

\textbf{Query-Loop Rules.}
We distinguish exact and near query loops. An exact query loop occurs when the
normalized query exactly matches an earlier query in the same trajectory. A
near query loop occurs when a non-identical query has normalized
character-trigram Jaccard similarity of at least $0.9$ and length of at least
eight characters. Exact matches take precedence over near matches. Invalid
actions and exact query loops are assigned severity $1.0$, while near query
loops are assigned severity $0.5$. Before query comparison, we apply
Unicode-NFKC normalization, case folding, zero-width character removal,
punctuation-to-space conversion, and whitespace collapsing.

\begin{table}[H]
	\centering
	\small
	\caption{Rule-based detector for undesirable behaviors.}
	\label{tab:detector_events}
	\setlength{\tabcolsep}{4pt}
	\renewcommand{\arraystretch}{1.15}
	\begin{tabular}{@{}p{0.22\linewidth}p{0.63\linewidth}c@{}}
		\toprule
		Undesirable behavior & Detection rule & Severity \\
		\midrule
		Invalid action
		& Malformed or impermissible action
		& $1.0$ \\
		\addlinespace
		\multirow{2}{*}{Query loop}
		& Exact repetition of an earlier normalized query
		& $1.0$ \\
		& Near repetition: trigram Jaccard $\geq 0.9$, length $\geq 8$
		& $0.5$ \\
		\bottomrule
	\end{tabular}
\end{table}

\textbf{Event Representation.}
The detector is implemented as a set of rule-based checks over generated
actions and tool interactions. Each detected event is represented as
\begin{equation}
	e=(I_e,d_e,q_e,J_e),
	\label{eq:app_event}
\end{equation}
where $I_e$ is the localized event span, $d_e$ is the behavior category,
$q_e\in(0,1]$ is its severity, and $J_e\subseteq I_e$ is the corresponding
action span. The event span $I_e$ determines where C$^3$ corrects credit,
whereas $J_e$ is used to estimate the policy confidence that controls how
strongly the correction is applied.

\textbf{Span Localization.}
For each detected event $e$, we use the complete turn responsible for the
event as its event span $I_e$, and the parsed tool or answer action within
that turn as its action span $J_e$. If the action cannot be parsed, we set
$J_e=I_e$. No-answer cases,
truncation, search-only termination, and strict-format-only failures are
tracked separately and do not generate detector events. When multiple events
overlap, each token receives the maximum severity among the events covering
it:
\begin{equation}
	q_i=\max_{e:i\in I_e}q_e,
	\qquad
	\mathcal{B}=\{i:q_i>0\}.
	\label{eq:app_severity}
\end{equation}
Tokens outside $\mathcal{B}$ are unaffected by C$^3$.

\subsection{Reward and Credit Construction}
\label{app:reward_credit}

\textbf{Outcome and Process Reward.}
Let $A$ indicate that
a final answer is present, $S$ that a search preceded it, $Y$ a relaxed exact
match, and $T$ truncation:
\begin{equation}
	R_{\mathrm{outcome}}=
	\begin{cases}
		0, & T=1,\\
		0.2AS+1.0YS, & T=0.
	\end{cases}
	\label{eq:environment_reward}
\end{equation}
Thus, a searched incorrect answer receives a small answer-presence reward,
while a searched correct answer receives the larger correctness reward. The
terminal-PPO arm adds the process penalty to this outcome reward at the
terminal position. The process-PPO arm instead places the same type of
penalty at the turn boundary where the undesirable behavior occurs.

\textbf{Credit Construction.}
The five training arms differ in both the signals used to construct credit and
the point at which detector information enters the optimization. Terminal PPO
adds the process penalty directly to the terminal outcome, after which PPO
computes GAE. Process-reward PPO instead assigns the penalty to the turn where
the undesirable behavior occurs and includes it in the return before GAE.
PPO+C$^3$ uses the outcome reward and leaves the base advantage construction
unchanged, applying a localized actor-side correction only after the base
advantages have been computed. GDPO constructs group-relative credit from
outcome, format, and process signals; GDPO+C$^3$ applies the same localized
post-estimation correction to that group-relative credit.

\textbf{Process-Reward PPO.}
For the process-reward PPO baseline, detected events are converted into
turn-level penalties. Invalid actions and exact query loops receive the full
penalty, while near query loops receive half strength. These penalties are
accumulated at the corresponding turn boundaries and incorporated into the
return before advantage estimation. In contrast, C$^3$ uses the same
behavioral localization but does not modify the task reward or return.

\textbf{GDPO Credit Construction.}
For GDPO, each four-response prompt group provides three sample-level axes:
outcome, format, and negative process cost. Each axis is normalized within
the group, the normalized axes are summed, and the resulting score is
normalized over the rollout batch:
\begin{equation}
	z_m^a=\frac{x_m^a-\mu_g^a}{\sigma_g^a+10^{-6}},
	\qquad
	A_m^{\mathrm{GDPO}}=
	\frac{\sum_a z_m^a-\mu_{\mathcal B}}
	{\sigma_{\mathcal B}+10^{-6}}.
	\label{eq:gdpo_advantage}
\end{equation}
For GDPO+C$^3$, this advantage is passed to C$^3$ without changing the
original GDPO normalization.

\subsection{Training and Evaluation Setup}
\label{app:training_eval}

\textbf{Models and Environments.}
For Search QA, we use Qwen3-8B with full-parameter updates and a fixed
Wiki-Fixed retrieval corpus~\citep{retrieve}. Retrieval uses
Qwen3-Embedding-8B representations with 4096 dimensions and exact
inner-product search with a top-$3$ cutoff. PPO and GDPO use the same corpus,
encoder, index type, and retrieval cutoff. WebShop uses Qwen3-4B with
full-parameter updates in the standard WebShop environment.

\textbf{Initialization and Training.}
Search QA RL runs are initialized from the corresponding Qwen3-8B SFT
checkpoint, while WebShop RL runs use the Qwen3-4B WebShop SFT checkpoint.
Within each comparison family, the initialization, task prompts, rollout
budgets, and evaluation definitions are kept fixed. For Search QA, PPO uses
GAE with batch whitening and a learned critic, while GDPO uses group-relative
credit without a learned critic. Each rollout group contains 64 prompts with
four responses per prompt. Generation uses temperature $1$, top-$p=1$, and
at most five turns. Models are trained with full-parameter bf16 updates using
Adam and gradient clipping. All paired baseline and C$^3$ arms use the same
configured RL training seed(s), with baseline and C$^3$ runs matched
seed-by-seed within each comparison family. This controls the paired training
randomness; evaluation-seed variation from auxiliary WebShop runs is reported
separately and should not be interpreted as additional RL training runs.

\textbf{Evaluation.}
Search QA evaluation uses greedy decoding with the common strict multi-hop
prompt. The evaluation suite contains seven benchmarks and 51,713 questions.
The evaluator extracts the final answer and computes normalized exact match,
with macro scores averaging equally across benchmarks. The multi-sample
results reported in the main text use temperature $0.8$ and top-$p=0.95$.
For WebShop, success is determined by the environment's final purchase
outcome, while task score is reported as the environment score.

\section{Additional Experiments}
\label{app:additional_experiments}
\subsection{Official T\texorpdfstring{$^2$}{2}PO WebShop Reproduction with C\texorpdfstring{$^3$}{3}}
\label{subsec:additional_t2po}

To further verify that C$^3$ functions as a \textit{\textbf{plug-in credit correction}},
we integrate it into the official T$^2$PO WebShop recipe. T$^2$PO~\citep{t2po}
is an uncertainty-aware exploration-control method for multi-turn agent
reinforcement learning that combines token-level and turn-level mechanisms to
improve exploration efficiency while preserving the underlying task reward and
policy optimization objective. Our reproduction follows the official T$^2$PO
training recipe and hyperparameter configuration, with C$^3$ inserted only
after the base T$^2$PO actor-side advantages are constructed.
\begin{table}[H]
	\centering
	\small
	\caption{Best validation performance.}
	\label{tab:additional_t2po}
	\setlength{\tabcolsep}{8pt}
	\renewcommand{\arraystretch}{1.08}
	\begin{tabular}{@{}lcccc@{}}
		\toprule
		Method & Success (\%) & Iter. & Task score (\%) & Iter. \\
		\midrule
		T2PO & 81.2 & 70 & 83.9 & 70 \\
		T2PO+C$^3$ & \textbf{82.2} & 120 & \textbf{86.0} & 130 \\
		\bottomrule
	\end{tabular}
\end{table}

The T$^2$PO and T$^2$PO+C$^3$ runs share the same SFT initialization, training
protocol, and rollout configuration. The only modification is that C$^3$ is
applied after the base T$^2$PO actor-side advantages are constructed, leaving
the underlying optimization pipeline unchanged.

\begin{figure}[H]
	\centering
	\includegraphics[width=0.98\linewidth]{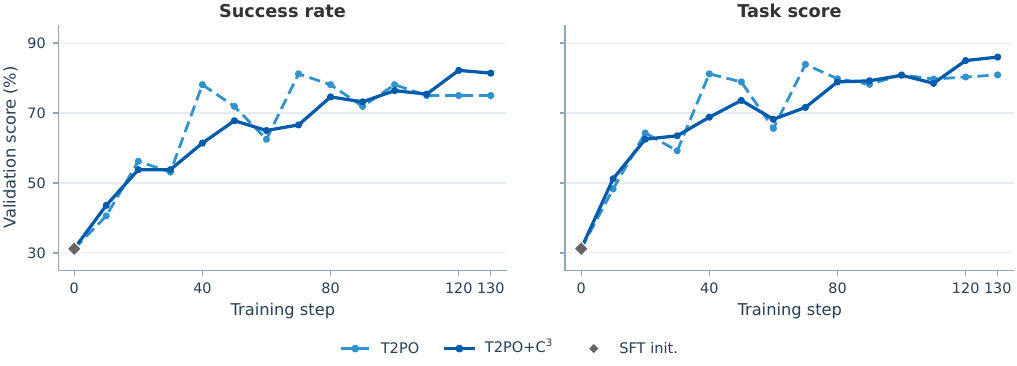}
	\caption{Validation trajectories for official T2PO and T2PO+C$^3$. The two
		panels show success rate and task score; gray diamonds mark the shared SFT
		initialization.}
	\label{fig:additional_t2po}
\end{figure}

Figure~\ref{fig:additional_t2po} compares the validation trajectories through
iteration 130. T$^2$PO+C$^3$ reaches a best success rate of \textbf{82.2\%},
compared with \textbf{81.2\%} for T$^2$PO, and improves the best task score
from \textbf{83.9\%} to \textbf{86.0\%}. 

\subsection{Seven-Benchmark Evaluation Dynamics}
\label{subsec:seven_benchmark_dynamics}

\begin{figure*}[t]
\centering
\includegraphics[width=0.99\textwidth]{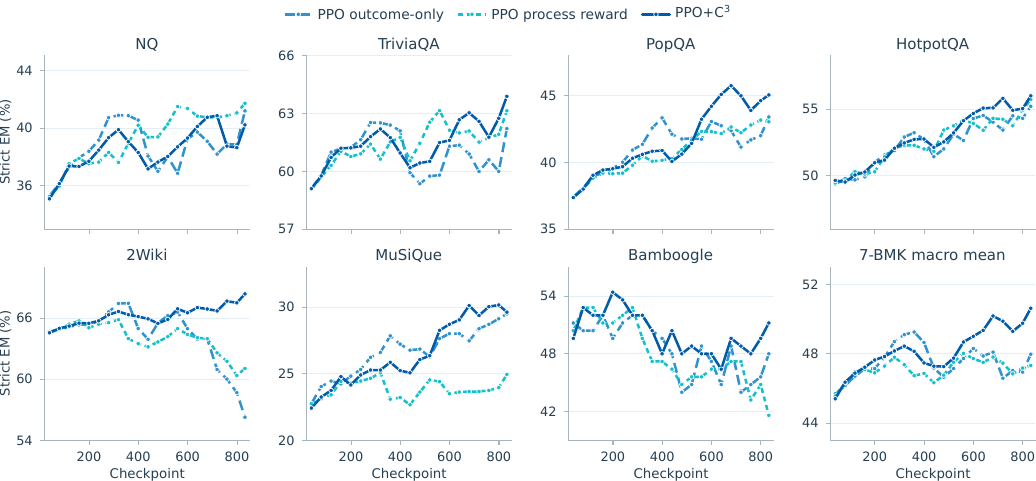}
\caption{Periodic strict-EM evaluation curves for the seven Search QA
benchmarks under the PPO family. The bottom-right panel shows the unweighted
7-BMK macro mean.}
\label{fig:appendix_7bmk_curves_ppo}
\end{figure*}

\begin{figure*}[h]
\centering
\includegraphics[width=0.99\textwidth]{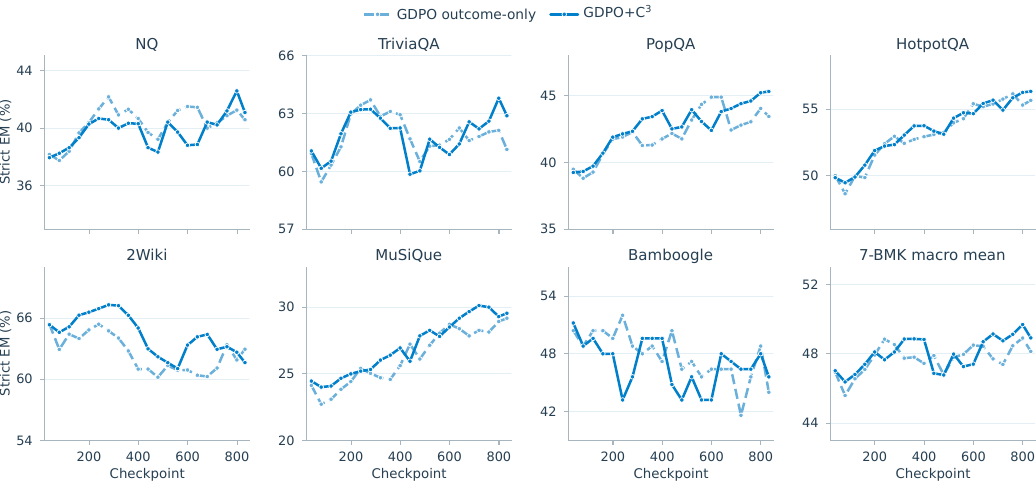}
\caption{Periodic strict-EM evaluation curves for the seven Search QA
benchmarks under the GDPO family. The bottom-right panel shows the unweighted
7-BMK macro mean.}
\label{fig:appendix_7bmk_curves_gdpo}
\end{figure*}

Figures~\ref{fig:appendix_7bmk_curves_ppo} and
\ref{fig:appendix_7bmk_curves_gdpo} decompose the macro evaluation curve into
the seven Search QA benchmarks, showing how each task contributes to the
overall trend under PPO and GDPO. Each point corresponds to the same periodic
evaluation checkpoint used in the main training dynamics.

The decomposition shows that the macro improvement arises from a distributed
effect across benchmarks rather than a single dominant task. Under PPO,
C$^3$ provides its clearest and most persistent gains on 2Wiki, with
additional improvements on PopQA and Bamboogle, while NQ remains broadly
stable. Under GDPO, C$^3$ improves NQ, TriviaQA, PopQA, HotpotQA, and MuSiQue
at the final checkpoint, whereas 2Wiki shows no corresponding gain in this
run. The benchmark-level curves also reveal that different tasks peak at
different stages of training, indicating that aggregate curves can obscure
meaningful intermediate behavior.

\subsection{Diagnosing Late WebShop Collapse}
\label{subsec:webshop_collapse}

The WebShop training curves reveal a pronounced late-stage collapse for
reward-side process PPO, while outcome-only GDPO exhibits a smaller late
degradation. To characterize this behavior, we analyze the same 500-task
WebShop benchmark used for periodic validation during training.

\begin{table}[H]
\centering
\footnotesize
\caption{Late-stage WebShop behavior on the 500-task validation benchmark.}
\label{tab:webshop_collapse_diagnostics}
\setlength{\tabcolsep}{4pt}
\renewcommand{\arraystretch}{1.12}
\begin{tabular}{@{}p{0.23\linewidth}p{0.16\linewidth}p{0.53\linewidth}@{}}
\toprule
Method & Success & Late-stage signature \\
\midrule
Process PPO
& $69.0\%\rightarrow11.2\%$
& Runaway reasoning, repetition, and severe truncation. \\
GDPO
& $68.8\%\rightarrow40.2\%$
& Shortened trajectories that terminate before purchase. \\
PPO+C$^3$
& $66.2\%\rightarrow60.8\%$
& No comparable collapse in the same window. \\
GDPO+C$^3$
& $71.4\%\rightarrow70.0\%$
& Stable performance through late training. \\
\bottomrule
\end{tabular}
\end{table}

To further diagnose the process-PPO collapse, Table~\ref{tab:webshop_turn_collapse}
summarizes turn-level statistics at three representative checkpoints.

\begin{table}[H]
\centering
\scriptsize
\caption{Turn-level signature of the process-PPO collapse.}
\label{tab:webshop_turn_collapse}
\setlength{\tabcolsep}{2.7pt}
\renewcommand{\arraystretch}{1.08}
\begin{tabular}{@{}p{0.28\linewidth}ccc p{0.27\linewidth}@{}}
\toprule
Statistic & 259 & 319 & 399 & Change (259$\rightarrow$399) \\
\midrule
Success rate (\%) & 69.0 & 26.6 & 11.2 & $-57.8$ pt \\
Truncated responses (\%) & 0.2 & 66.6 & 86.2 & $+86.0$ pt \\
Trailing \texttt{<think>} median (chars) & 328 & 7{,}962 & 8{,}432 & $\approx26\times$ \\
Generated tokens / trajectory & 1{,}043 & 2{,}536 & 2{,}667 & $+155\%$ \\
Completed turns / trajectory & 18.9 & 11.2 & 7.7 & $-59\%$ \\
Mean assistant chars / turn & 951 & 1{,}245 & 1{,}354 & $+42\%$ \\
Total trajectory chars & 17{,}965 & 18{,}739 & 16{,}648 & $-7\%$ \\
\bottomrule
\end{tabular}
\end{table}

The collapse is characterized by a runaway final reasoning block rather than
a uniformly longer interaction. The median trailing
\texttt{<think>} segment expands from 328 to 8,432 characters, while the
number of completed turns falls from 18.9 to 7.7. Generated length increases
because the unfinished reasoning block consumes the response budget, whereas
the overall trajectory becomes shorter because later turns never occur.
Search frequency and detector-targeted invalid actions remain comparatively
stable, indicating that the collapse reflects a termination failure in the
final reasoning stage rather than an increase in the targeted undesirable
behaviors.

By contrast, both C$^3$ variants avoid a comparable late-stage collapse over
the same training window. Since C$^3$ modifies only localized actor-side
credit after advantage construction while leaving reward and return estimation
unchanged, the observed stability is consistent with its localized
intervention mechanism.

\subsection{How Does C\texorpdfstring{$^3$}{3} Affect Interaction Behavior?}
\label{subsec:analysis_retrieval}

\begin{figure}[H]
	\centering
	\includegraphics[width=0.98\linewidth]{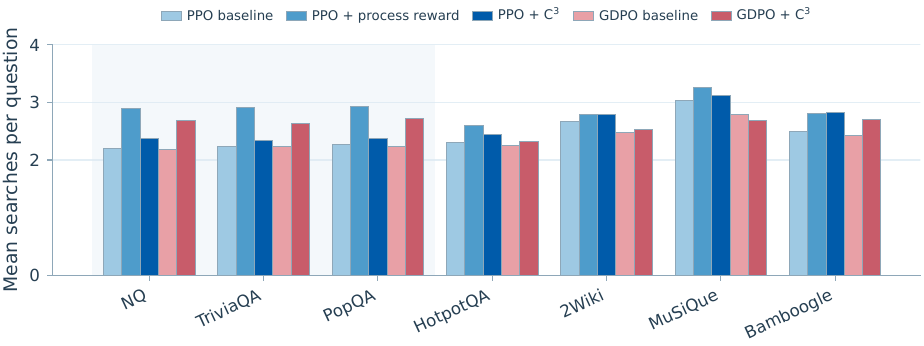}
	\caption{Mean searches across seven benchmarks.
		Shaded columns denotes single-hop tasks.}
	\label{fig:per_benchmark}
\end{figure}

C$^3$ operates by correcting local credit assigned to detected undesirable
behaviors rather than directly controlling the number of tool interactions.
Its effect on interaction behavior may therefore depend on both the
underlying optimization paradigm and the requirements of the task. We
examine this effect by comparing the mean number of searches per question
across the PPO and GDPO families. Figure~\ref{fig:per_benchmark} reports the
absolute search counts for all seven benchmarks, with single-hop and
multi-hop tasks shown separately.

Under \textbf{PPO}, C$^3$ changes retrieval behavior differently across task
types. On the three single-hop benchmarks, C$^3$ reduces the mean number of
searches by $0.09$--$0.17$ per question relative to outcome-only PPO. In
contrast, C$^3$ increases the number of searches on the multi-hop
benchmarks. This pattern suggests that C$^3$ does not simply push the agent
toward shorter or longer interaction trajectories. Rather, after correcting
credit on detected undesirable behaviors, the resulting policy can allocate
retrieval differently depending on the information requirements of the
task. In particular, reducing unnecessary interactions on simpler tasks
does not preclude retaining additional retrieval on tasks that require
multi-hop information gathering.

The behavior under \textbf{GDPO} is different. Compared with outcome-only
GDPO, GDPO+C$^3$ generally increases the number of searches across the
benchmarks. This difference from PPO highlights that the behavioral effect
of the same local credit correction is not invariant to the underlying
optimization paradigm. PPO and GDPO construct token-level credit
differently, so correcting the same type of local positive credit can lead
to different changes in the resulting interaction policy. Importantly, the
increase in search frequency under GDPO should not be interpreted as C$^3$
directly encouraging more retrieval, since retrieval frequency is not an
explicit optimization target of the method.

Taken together, these results show that \textit{\textbf{C$^3$ does not
		impose a fixed preference for shorter or longer interaction trajectories}}.
Instead, its behavioral effect emerges indirectly through local credit
correction and varies with both the optimization paradigm and the task.
This distinction is important because it suggests that the performance
gains from C$^3$ should not be understood simply in terms of performing
more or fewer tool interactions. C$^3$ changes which local behaviors receive
policy credit, while the resulting interaction allocation is determined by
the broader optimization process and task requirements.

% The case studies use compact colored cards so that a reader can scan the
% task, the behavioral fork, and the interpretation without reading raw logs.
\definecolor{CaseBlueBg}{HTML}{F4F8FC}
\definecolor{CaseBlueFrame}{HTML}{8AA8C0}
\definecolor{CaseGreenBg}{HTML}{F3F9F5}
\definecolor{CaseGreenFrame}{HTML}{8EB79D}
\definecolor{CaseRoseBg}{HTML}{FCF4F5}
\definecolor{CaseRoseFrame}{HTML}{C9959D}
\definecolor{CaseGoldBg}{HTML}{FFF9EA}
\definecolor{CaseGoldFrame}{HTML}{D0B36A}
\definecolor{CaseInk}{HTML}{33404A}

\newtcolorbox{casecard}[2][]{
  enhanced,
  colback=#2, colframe=#2!65!black,
  boxrule=0.55pt, leftrule=2.2pt, arc=1.1mm,
  title={#1}, fonttitle=\bfseries\small, coltitle=CaseInk,
  attach boxed title to top left={xshift=2mm,yshift*=-\tcboxedtitleheight/2},
  boxed title style={colback=#2!78!black, colframe=#2!78!black,
    boxrule=0pt, arc=0.8mm},
  top=2.8mm, bottom=1.7mm, left=2.5mm, right=2.5mm,
  before skip=2.2mm, after skip=2.2mm
}

\newtcolorbox{caseprompt}{
  enhanced, colback=white, colframe=CaseBlueFrame!55,
  boxrule=0.35pt, leftrule=1.4pt, arc=0.7mm,
  top=1.2mm, bottom=0.9mm, left=2mm, right=2mm,
  before skip=1.1mm, after skip=1.4mm
}

\newtcblisting{promptpanel}[1][]{
  enhanced, breakable, listing only,
  colback=white, colframe=CaseBlueFrame!55,
  boxrule=0.35pt, leftrule=1.2pt, arc=0.7mm,
  title={#1}, fonttitle=\bfseries\footnotesize, coltitle=CaseInk,
  boxed title style={colback=CaseBlueBg, colframe=CaseBlueFrame!55,
    boxrule=0.35pt, arc=0.5mm},
  attach boxed title to top left={xshift=1.5mm,yshift*=-\tcboxedtitleheight/2},
  top=1.4mm, bottom=0.9mm, left=1.7mm, right=1.7mm,
  before skip=1.1mm, after skip=1.4mm,
  listing options={basicstyle=\ttfamily\fontsize{6.5}{7.1}\selectfont,
    columns=fullflexible, keepspaces=true, breaklines=true,
    breakatwhitespace=false, showstringspaces=false}
}

\newtcblisting{casepanel}[1][]{
  enhanced, breakable, listing only,
  colback=white, colframe=CaseBlueFrame!55,
  boxrule=0.35pt, leftrule=1.2pt, arc=0.7mm,
  title={#1}, fonttitle=\bfseries\footnotesize, coltitle=CaseInk,
  boxed title style={colback=CaseBlueBg, colframe=CaseBlueFrame!55,
    boxrule=0.35pt, arc=0.5mm},
  attach boxed title to top left={xshift=1.5mm,yshift*=-\tcboxedtitleheight/2},
  top=1.4mm, bottom=0.9mm, left=1.7mm, right=1.7mm,
  before skip=1.1mm, after skip=1.4mm,
  listing options={basicstyle=\ttfamily\fontsize{7.2}{8.4}\selectfont,
    columns=fullflexible, keepspaces=true, breaklines=true,
    breakatwhitespace=false, showstringspaces=false}
}

\newtcolorbox{caseframe}[2][] {
  enhanced,
  colback=#2!4, colframe=#2,
  boxrule=0.6pt, arc=1.2mm,
  title={#1}, fonttitle=\bfseries\footnotesize, coltitle=CaseInk,
  attach boxed title to top left={xshift=2mm,yshift*=-\tcboxedtitleheight/2},
  boxed title style={colback=#2!20, colframe=#2, boxrule=0.4pt, arc=0.8mm},
  left=2mm, right=2mm, top=2.2mm, bottom=1.2mm,
  before skip=0.6mm, after skip=0.6mm
}

\newtcolorbox{caseturnbox}[1][] {
  enhanced,
  colback=white, colframe=CaseBlueFrame!55,
  boxrule=0.35pt, arc=0.8mm,
  left=1.3mm, right=1.3mm, top=0.7mm, bottom=0.7mm,
  before skip=0.35mm, after skip=0.35mm,
  title={#1}, fonttitle=\bfseries\scriptsize, coltitle=CaseInk,
  attach boxed title to top left={xshift=1.3mm,yshift*=-\tcboxedtitleheight/2},
  boxed title style={colback=CaseBlueBg, colframe=CaseBlueFrame!55,
    boxrule=0.3pt, arc=0.5mm}
}

\newtcolorbox{casefinalbox}[1][] {
  enhanced,
  colback=CaseGreenBg, colframe=CaseGreenFrame,
  boxrule=0.55pt, arc=0.8mm,
  left=1.3mm, right=1.3mm, top=0.7mm, bottom=0.7mm,
  before skip=0.35mm, after skip=0.35mm,
  title={#1}, fonttitle=\bfseries\scriptsize, coltitle=CaseInk,
  attach boxed title to top left={xshift=1.3mm,yshift*=-\tcboxedtitleheight/2},
  boxed title style={colback=CaseGreenBg, colframe=CaseGreenFrame,
    boxrule=0.3pt, arc=0.5mm}
}

\newcommand{\caseaction}[1]{\texttt{\scriptsize #1}}
\newcommand{\casegood}{\textcolor{CaseGreenFrame}{\textbf{Correct}}}
\newcommand{\casebad}{\textcolor{CaseRoseFrame}{\textbf{Failure}}}

\section{Case Studies}
\label{app:case_studies}

\subsection{Prompt Interfaces}
\label{app:prompt_interfaces}

The panels below reproduce the complete prompt interfaces used in the
experiments. Braced fields are filled by the task runner; the concrete values
used by the saved trajectories are shown separately in the case studies. Panel
boundaries are typographic only and do not indicate omitted prompt content.

\paragraph{Initial Search QA prompt.}
At the initial turn, the runner sends one system instruction followed by the
user question. The system instruction is split into the panels below solely to
make the appendix readable; the panels preserve the original order and wording.

\begin{promptpanel}[Role and objective]
You are solving a retrieval-based question answering task. Some questions can be answered directly, while others require one or more searches to gather evidence.

Your goal is to answer correctly, not to minimize the number of searches. Search whenever the available evidence is insufficient to directly support the answer.
\end{promptpanel}

\begin{promptpanel}[Decision process]
## Decision Process

For every turn:

### Step 1. Understand the question.

Determine what information is required to answer the original question.

The question may be:

- Direct: the answer is already known from the retrieved evidence.
- Single-hop: one search is sufficient.
- Multi-hop: multiple searches are required because the first search only reveals an intermediate entity, title, date, location, or relation.

### Step 2. Examine the available evidence.

Ask yourself:

Does the available evidence directly answer the original question?

If yes, output <answer>...</answer>.

If no, identify the most important missing fact and search for it. Output <search>...</search>.

Do not answer from memory when a targeted search can verify the claim.
\end{promptpanel}

\begin{promptpanel}[Search strategy]
## Search Strategy

Each search should retrieve one missing fact.

Good search queries usually consist of:
- entity + property
- entity + relation
- title + attribute

Examples:
Titanic director
Alan Turing birthplace
The Beatles members
Mount Everest elevation

For multi-hop questions:
- The first search may only identify an intermediate entity.
- Use the newly discovered entity to construct the next search.
- Continue until the retrieved evidence directly supports the final answer.

Do not repeat the same search query. If a result is ambiguous, incomplete, or off-topic, issue a more specific search instead of guessing.
\end{promptpanel}

\begin{promptpanel}[Evidence sufficiency]
## Evidence Sufficiency

Evidence is sufficient only if it directly supports the final answer to the original question.

The following alone are not sufficient:
- an intermediate entity
- a related title
- a date
- a location
- a relationship
- a weak or indirect match

If only intermediate information has been retrieved, search again.
\end{promptpanel}

\begin{promptpanel}[Output format]
## Output Format

Every assistant turn must contain exactly one action.

Either
<think>...</think>
<search>...</search>

or
<think>...</think>
<answer>...</answer>

Never output both <search> and <answer> in the same turn.
Never output text outside these tags.
Always begin with <think>.
\end{promptpanel}

\begin{promptpanel}[Few-shot examples]
## Examples

### Example 1 (single-hop)
Question: What is the capital of France?
<think>I need to verify the capital of France.</think>
<search>France capital</search>

Later:
<think>The retrieved evidence directly states the capital.</think>
<answer>Paris</answer>

### Example 2 (multi-hop)
Question: Who directed the movie starring the actor who played Jack Dawson?
<think>I first need to identify the movie associated with Jack Dawson.</think>
<search>Jack Dawson movie</search>

Later:
<think>The movie has been identified, but the director is still unknown.</think>
<search>Titanic director</search>

Later:
<think>The retrieved evidence directly answers the question.</think>
<answer>James Cameron</answer>

### Example 3 (ambiguous search)
<think>The previous search returned multiple unrelated entities. I should make the query more specific.</think>
<search>Mercury planet orbital period</search>
\end{promptpanel}

\begin{promptpanel}[Rules and user message]
## Rules
1. Answer only when the retrieved evidence directly supports the original question.
2. Search whenever an important fact is still missing.
3. Do not infer missing facts without supporting evidence.
4. Do not repeat previous search queries.
5. If uncertain, search rather than guess.
6. Produce exactly one action per turn.
7. Invalid output format may be rejected.

--- user message ---
Question: {question}
\end{promptpanel}

After each search, the runner appends the retrieved passages and applies the
following compact check before the next assistant turn:

\begin{promptpanel}[Search observation guidance]
EVIDENCE_CHECK: If the evidence answers the question, output one <think>...</think> followed by <answer>short final answer</answer>; otherwise output one <think>...</think> followed by <search>one targeted query</search>. Do not repeat queries or write outside the tags.
\end{promptpanel}

\paragraph{Initial WebShop prompt.}
At the initial turn, the runner instantiates the task description, current
observation, and admissible actions in the instruction below. The complete
interface is split into ordered panels for readability; at runtime the panels
are concatenated without changing their content or order.

\begin{promptpanel}[Task and checklist]
You are an expert autonomous agent operating in the WebShop e-commerce environment.

Your task is to complete this shopping request:
{task_description}

Maintain a concise checklist of the explicit requirements:
product category, required attributes or features, price limit, and required
options such as color, size, flavor, quantity, or capacity.

Choose the admissible action that makes the most direct progress toward a valid purchase.
\end{promptpanel}

\begin{promptpanel}[Decision strategy: search and filter]
Decision strategy:

1. SEARCH
   If no suitable candidate is visible, use a concise query containing the
   most distinctive product requirements.

2. FILTER
   On a results page, inspect visible titles, attributes, and prices. Open
   only a plausible candidate. Do not click an irrelevant product merely
   because it is visible.
\end{promptpanel}

\begin{promptpanel}[Decision strategy: verify and select]
3. VERIFY
   On a product page, verify the explicit requirements using the visible
   title, price, and page information. Use Description or Features only
   when necessary to resolve a missing requirement.

4. SELECT OPTIONS
   If a required option is visible and unselected, select it before buying.
\end{promptpanel}

\begin{promptpanel}[Decision strategy: purchase and recover]
5. PURCHASE
   When all explicit requirements are satisfied or sufficiently verified,
   all required options are selected, and Buy Now is admissible, click Buy Now.

6. RECOVER
   If the candidate fails an explicit requirement, return to search results
   and use a refined query, another visible candidate, or the next page.
   Do not repeat the same failed query or candidate.

7. AVOID LOOPS
   Do not repeatedly inspect the same information or navigate without
   gaining evidence relevant to the shopping request.
\end{promptpanel}

\begin{promptpanel}[Action rules]
Action rules:
- search[<your query>] is a template; replace it with a concrete query.
- Never output a literal placeholder such as <your query> or concise keywords.
- For click actions, copy the exact admissible action.
- Prefer search[...] over click[search] when a search action is available.
- Choose exactly one admissible action.
- Keep the reasoning concise: one to three sentences.

Your reasoning MUST be enclosed within <think> </think> tags.
Your selected action MUST be enclosed within <action> </action> tags.
Do not output any text outside these two tags.
\end{promptpanel}

\begin{promptpanel}[Initial observation and output]
Current observation:
{current_observation}

Admissible actions:
[
{available_actions}
]

Return exactly:
<think>One to three concise sentences of reasoning.</think>
<action>Exactly one admissible action.</action>
\end{promptpanel}

For later WebShop turns, the runner appends only the new observation and
admissible-action block; the task description and decision strategy are not
repeated:

\begin{promptpanel}[Later-turn append block]
Current observation:
{current_observation}
Admissible actions:
[
{available_actions}
]
Return exactly:
<think>One to three concise sentences of reasoning.</think>
<action>Exactly one admissible action.</action>
\end{promptpanel}

\begin{figure*}[t]
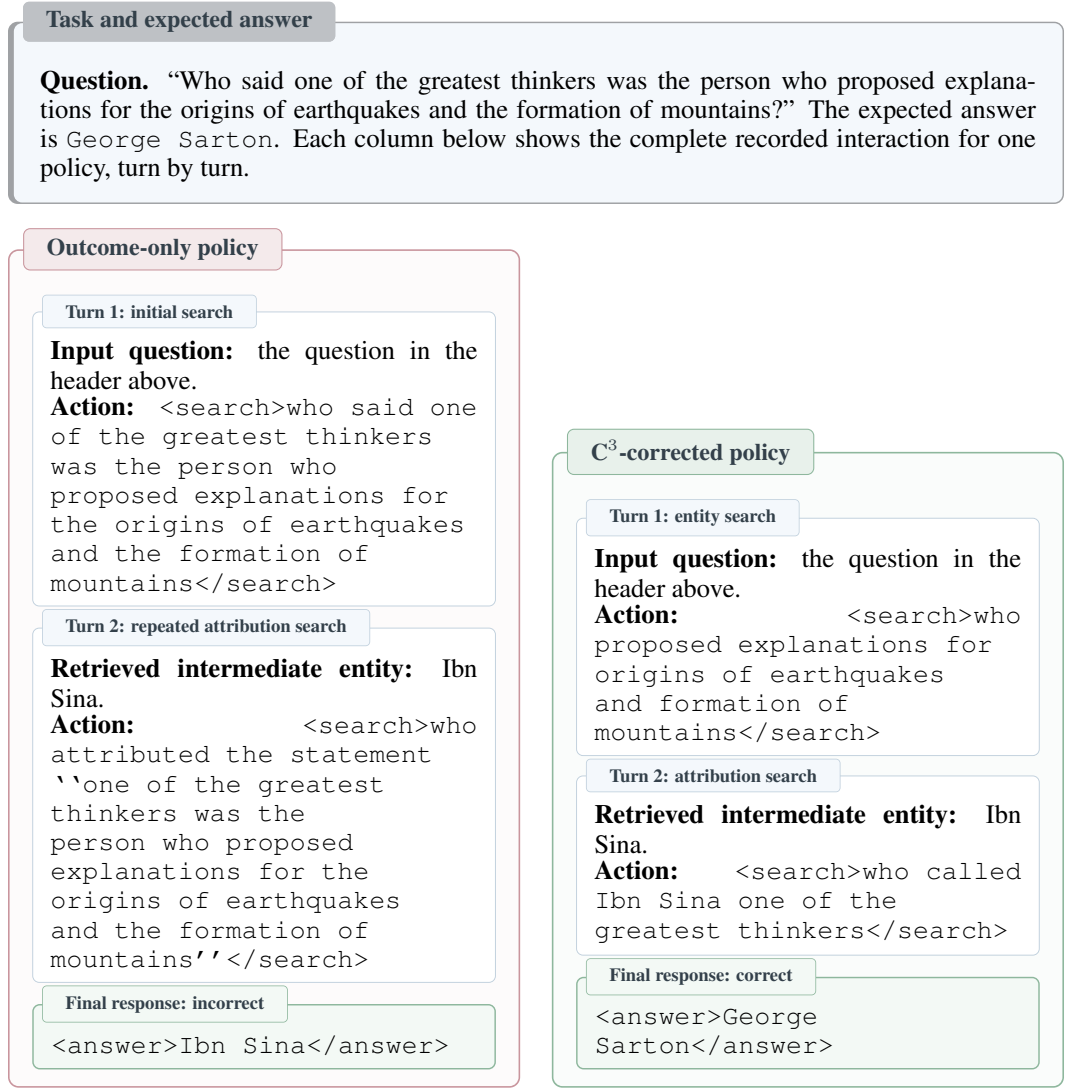

\subsection{Search QA: Complete Multi-hop Trajectory}
\label{app:case_search_complete}
\centering
\begin{casecard}[Task and expected answer]{CaseBlueBg}
\textbf{Question.} ``Who said one of the greatest thinkers was the person who
proposed explanations for the origins of earthquakes and the formation of
mountains?'' The expected answer is \texttt{George Sarton}. Each column below
shows the complete recorded interaction for one policy, turn by turn.
\end{casecard}

\vspace{1mm}
\noindent
\begin{minipage}[t]{0.485\textwidth}
\begin{caseframe}[Outcome-only policy]{CaseRoseFrame}
\begin{caseturnbox}[Turn 1: initial search]
\textbf{Input question:} the question in the header above.\\[-1pt]
\textbf{Action:} \texttt{<search>who said one of the greatest thinkers was the person who proposed explanations for the origins of earthquakes and the formation of mountains</search>}
\end{caseturnbox}
\begin{caseturnbox}[Turn 2: repeated attribution search]
\textbf{Retrieved intermediate entity:} Ibn Sina.\\[-1pt]
\textbf{Action:} \texttt{<search>who attributed the statement ``one of the greatest thinkers was the person who proposed explanations for the origins of earthquakes and the formation of mountains''</search>}
\end{caseturnbox}
\begin{casefinalbox}[Final response: incorrect]
\texttt{<answer>Ibn Sina</answer>}
\end{casefinalbox}
\end{caseframe}
\end{minipage}%
\hfill
\begin{minipage}[t]{0.485\textwidth}
\begin{caseframe}[C$^3$-corrected policy]{CaseGreenFrame}
\begin{caseturnbox}[Turn 1: entity search]
\textbf{Input question:} the question in the header above.\\[-1pt]
\textbf{Action:} \texttt{<search>who proposed explanations for origins of earthquakes and formation of mountains</search>}
\end{caseturnbox}
\begin{caseturnbox}[Turn 2: attribution search]
\textbf{Retrieved intermediate entity:} Ibn Sina.\\[-1pt]
\textbf{Action:} \texttt{<search>who called Ibn Sina one of the greatest thinkers</search>}
\end{caseturnbox}
\begin{casefinalbox}[Final response: correct]
\texttt{<answer>George Sarton</answer>}
\end{casefinalbox}
\end{caseframe}
\end{minipage}

\caption{Complete Search QA case presentation. The outcome-only policy repeats
the original relation, while the C$^3$-corrected policy separates entity
identification from attribution and returns the requested person.}
\label{fig:case_search_complete}
\raggedright\par\smallskip\noindent
\textbf{Interpretation.} The useful second turn is a new evidence bridge rather
than a longer trajectory. The side-by-side presentation makes the local fork
visible: the baseline repeats the relation and answers with the intermediate
entity, while the corrected policy verifies the attribution before answering.
\end{figure*}

\begin{figure*}[t]
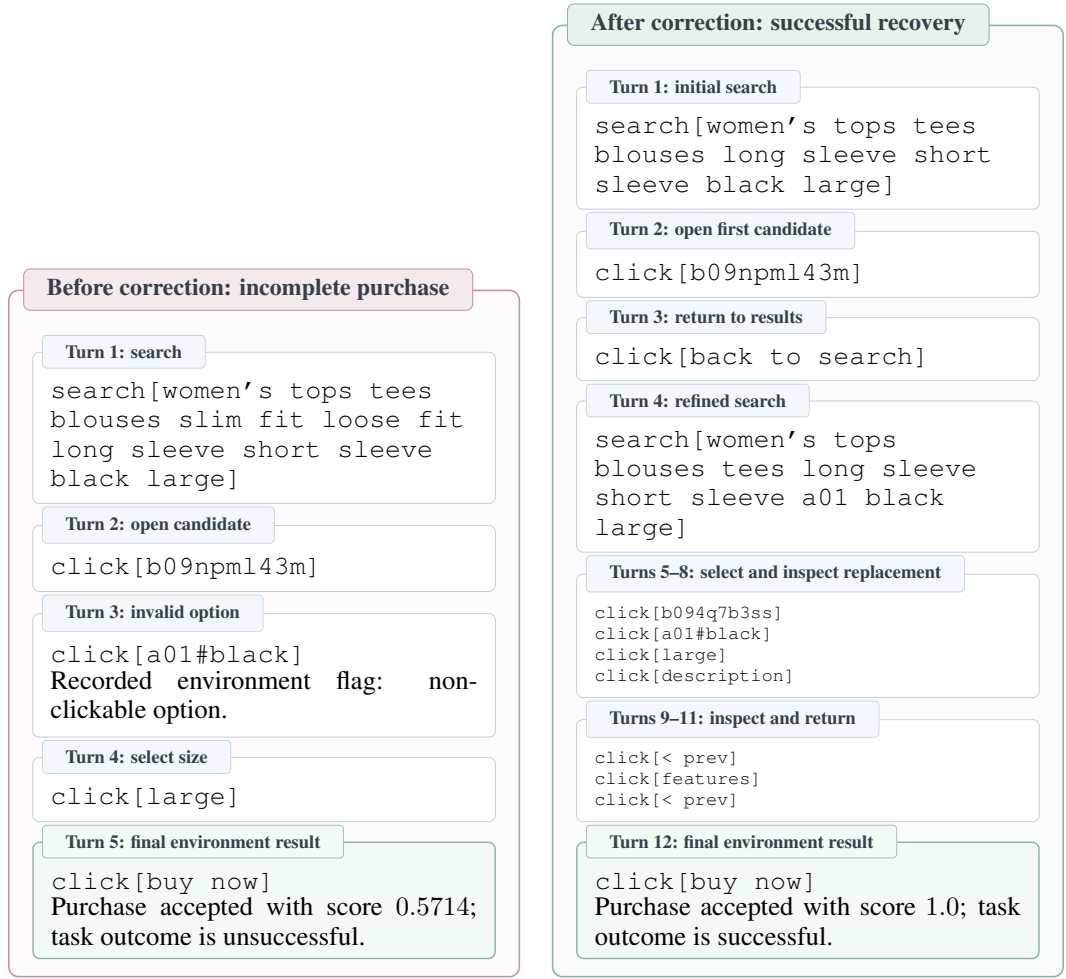

\subsection{WebShop: Complete Recovery Before Purchase}
\label{app:case_webshop_complete}
\centering
\begin{casecard}[Task and environment]{CaseBlueBg}
\textbf{Task.} ``Find me loose fit women's tops, tees \& blouses with short
sleeve, long sleeve with color \texttt{a01\#black}, and size large, and price
lower than 40.00 dollars.'' The two columns show the complete before/after
interaction for this same task, including every recorded environment action.
\end{casecard}

\vspace{1mm}
\noindent
\begin{minipage}[t]{0.485\textwidth}
\begin{caseframe}[Before correction: incomplete purchase]{CaseRoseFrame}
\begin{caseturnbox}[Turn 1: search]
\texttt{search[women's tops tees blouses slim fit loose fit long sleeve short sleeve black large]}
\end{caseturnbox}
\begin{caseturnbox}[Turn 2: open candidate]
\texttt{click[b09npml43m]}
\end{caseturnbox}
\begin{caseturnbox}[Turn 3: invalid option]
\texttt{click[a01\#black]}\\[-1pt]
Recorded environment flag: non-clickable option.
\end{caseturnbox}
\begin{caseturnbox}[Turn 4: select size]
\texttt{click[large]}
\end{caseturnbox}
\begin{casefinalbox}[Turn 5: final environment result]
\texttt{click[buy now]}\\[-1pt]
Purchase accepted with score $0.5714$; task outcome is unsuccessful.
\end{casefinalbox}
\end{caseframe}
\end{minipage}%
\hfill
\begin{minipage}[t]{0.485\textwidth}
\begin{caseframe}[After correction: successful recovery]{CaseGreenFrame}
\begin{caseturnbox}[Turn 1: initial search]
\texttt{search[women's tops tees blouses long sleeve short sleeve black large]}
\end{caseturnbox}
\begin{caseturnbox}[Turn 2: open first candidate]
\texttt{click[b09npml43m]}
\end{caseturnbox}
\begin{caseturnbox}[Turn 3: return to results]
\texttt{click[back to search]}
\end{caseturnbox}
\begin{caseturnbox}[Turn 4: refined search]
\texttt{search[women's tops blouses tees long sleeve short sleeve a01 black large]}
\end{caseturnbox}
\begin{caseturnbox}[Turns 5--8: select and inspect replacement]
{\ttfamily\scriptsize
click[b094q7b3ss]\par
click[a01\#black]\par
click[large]\par
click[description]\par}
\end{caseturnbox}
\begin{caseturnbox}[Turns 9--11: inspect and return]
{\ttfamily\scriptsize
click[< prev]\par
click[features]\par
click[< prev]\par}
\end{caseturnbox}
\begin{casefinalbox}[Turn 12: final environment result]
\texttt{click[buy now]}\\[-1pt]
Purchase accepted with score $1.0$; task outcome is successful.
\end{casefinalbox}
\end{caseframe}
\end{minipage}

\caption{Complete WebShop case presentation. The later policy returns to search
after the first candidate fails an explicit requirement, refines the query,
and verifies the replacement before purchase.}
\label{fig:case_webshop_complete}
\raggedright\par\smallskip\noindent
\textbf{Interpretation.} The important behavioral change is recovery from a
bad candidate, not merely the number of actions. The full turn layout exposes
the invalid option, the return-to-search decision, and the final purchase.
\end{figure*}

\begin{figure*}[t]
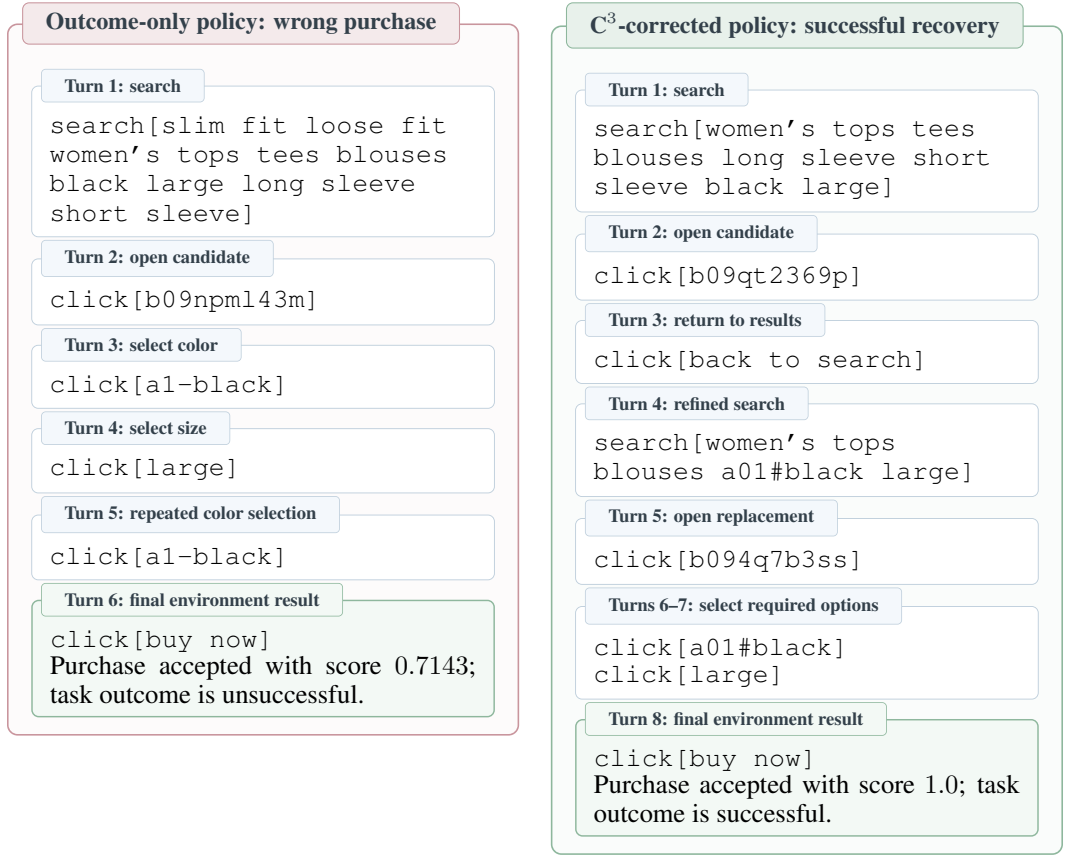

\subsection{WebShop: Complete Failure and Recovery Contrast}
\label{app:case_webshop_contrast}
\centering
\begin{casecard}[One task and two complete trajectories]{CaseRoseBg}
The same shopping request is evaluated under an outcome-only policy and a
C$^3$-corrected policy. The full turn sequence makes the local decision fork
visible: one policy purchases after repeatedly selecting the wrong option,
whereas the other changes candidate before purchase.
\end{casecard}

\vspace{1mm}
\noindent
\begin{minipage}[t]{0.485\textwidth}
\vspace{0pt}
\begin{caseframe}[Outcome-only policy: wrong purchase]{CaseRoseFrame}
\begin{caseturnbox}[Turn 1: search]
\texttt{search[slim fit loose fit women's tops tees blouses black large long sleeve short sleeve]}
\end{caseturnbox}
\begin{caseturnbox}[Turn 2: open candidate]
\texttt{click[b09npml43m]}
\end{caseturnbox}
\begin{caseturnbox}[Turn 3: select color]
\texttt{click[a1-black]}
\end{caseturnbox}
\begin{caseturnbox}[Turn 4: select size]
\texttt{click[large]}
\end{caseturnbox}
\begin{caseturnbox}[Turn 5: repeated color selection]
\texttt{click[a1-black]}
\end{caseturnbox}
\begin{casefinalbox}[Turn 6: final environment result]
\texttt{click[buy now]}\\[-1pt]
Purchase accepted with score $0.7143$; task outcome is unsuccessful.
\end{casefinalbox}
\end{caseframe}
\end{minipage}%
\hfill
\begin{minipage}[t]{0.485\textwidth}
\vspace{0pt}
\begin{caseframe}[C$^3$-corrected policy: successful recovery]{CaseGreenFrame}
\begin{caseturnbox}[Turn 1: search]
\texttt{search[women's tops tees blouses long sleeve short sleeve black large]}
\end{caseturnbox}
\begin{caseturnbox}[Turn 2: open candidate]
\texttt{click[b09qt2369p]}
\end{caseturnbox}
\begin{caseturnbox}[Turn 3: return to results]
\texttt{click[back to search]}
\end{caseturnbox}
\begin{caseturnbox}[Turn 4: refined search]
\texttt{search[women's tops blouses a01\#black large]}
\end{caseturnbox}
\begin{caseturnbox}[Turn 5: open replacement]
\texttt{click[b094q7b3ss]}
\end{caseturnbox}
\begin{caseturnbox}[Turns 6--7: select required options]
\texttt{click[a01\#black]}\\[-1pt]
\texttt{click[large]}
\end{caseturnbox}
\begin{casefinalbox}[Turn 8: final environment result]
\texttt{click[buy now]}\\[-1pt]
Purchase accepted with score $1.0$; task outcome is successful.
\end{casefinalbox}
\end{caseframe}
\end{minipage}

\caption{Complete WebShop failure/recovery case. The outcome-only trajectory
keeps the first candidate and repeats an invalid option; the C$^3$ trajectory
returns to search, changes candidate, and completes the requested purchase.}
\label{fig:case_webshop_contrast}
\raggedright\par\smallskip\noindent
\textbf{Interpretation.} The contrast is about which local action receives
credit, not about making the trajectory uniformly longer or shorter. The
corrected policy changes candidate after the first result and reaches the
required option set before buying.
\end{figure*}

\end{document}